\documentclass[letterpaper, 10 pt, conference]{ieeeconf}  

\IEEEoverridecommandlockouts                              

\UseRawInputEncoding

\usepackage{graphics} 
\usepackage{epsfig} 
\usepackage{times} 
\usepackage{amsmath} 
\usepackage{amssymb}  
\usepackage{eurosym,fancyhdr,CJK,multicol,graphics,indentfirst,color,bm,upgreek,booktabs,graphicx}

\usepackage{multirow}
\usepackage{makecell}
\usepackage{pifont}

\usepackage{flushend}
\usepackage{cite}
\usepackage[colorlinks,breaklinks]{hyperref}

\title{\LARGE \bf Semantic-aware Texture-Structure Feature Collaboration for Underwater Image Enhancement}

\author{ Di Wang$^1$, Long Ma$^1$, Risheng Liu$^2$ and Xin Fan$^{2,*}$

\thanks{*Corresponding Author: {\tt xin.fan@dlut.edu.cn}}

\thanks{$^{1}$School of Software Technology, Dalian University of Technology, Dalian, 116024, China.
}
\thanks{$^{2}$DUT-RU International School of Information Science \& Engineering, Dalian University of Technology, Dalian, 116024, China.
}%
\thanks{This work was supported by the National Key R\&D Program of
	China (2020YFB1313503), the National Natural Science Foundation of China
	(Nos. 61922019, 61733002, and 62027826), and LiaoNing Revitalization
	Talents Program (XLYC1807088).}
}

\begin{document}

\maketitle
\thispagestyle{empty}
\pagestyle{empty}

\begin{abstract}
	
	Underwater image enhancement has become an attractive topic as a significant technology in marine engineering and aquatic robotics.
	However, the limited number of datasets and imperfect hand-crafted ground truth weaken its robustness to unseen scenarios, and hamper the application to high-level vision tasks. 
	To address the above limitations, we develop an efficient and compact enhancement network in collaboration with a high-level semantic-aware pretrained model, aiming to exploit its hierarchical feature representation as an auxiliary for the low-level underwater image enhancement.
	%
	%
	Specifically, we tend to characterize the shallow layer features as textures while the deep layer features as structures in the semantic-aware model, and propose a multi-path Contextual Feature Refinement Module (CFRM) to refine features in multiple scales and model the correlation between different features.
	%
	%
	In addition, a feature dominative network is devised to perform channel-wise modulation on the aggregated texture and structure features for the adaptation to different feature patterns of the enhancement network. 
	Extensive experiments on benchmarks demonstrate that the proposed algorithm achieves more appealing results and outperforms state-of-the-art methods by large margins.
	We also apply the proposed algorithm to the underwater salient object detection task to reveal the favorable semantic-aware ability for high-level vision tasks.
	This code is available at~\href{https://github.com/wdhudiekou/STSC}{STSC}.
	
\end{abstract}

\section{INTRODUCTION}

Underwater image enhancement is a practical but challenging technology in the field of underwater vision, which is widely contributed to many applications such as aquatic robotics~\cite{James_robot_2021}, underwater path planning~\cite{Cagara_ICRA_path}, and underwater object real-time tracking~\cite{Langis_ICRA_track}, etc. 
Over the past few decades, a series of underwater enhancement methods also have been explored, ranging from traditional model-free methods~\cite{Fu_retinex, Fu_TSA, Li_OCM, Ancuti_EUIVF} to physical model-based methods~\cite{McGlamery1980ACM,Uplavikar_AIO,Li_Jaffe,Peng_Jaffe,Drews_UDCP}. 
%
%

\begin{figure}[t]
	\footnotesize
	\begin{center}
		\begin{tabular}{cc}
			\includegraphics[width = 0.38\linewidth]{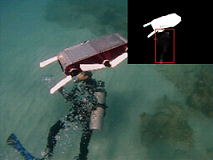} &
			\hspace{-1.6mm}
			\includegraphics[width = 0.38\linewidth]{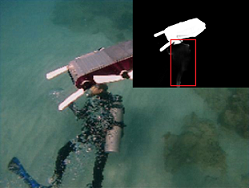}\\
			(a) WaterNet~\cite{Li_waternet} & (b) UWCNN~\cite{Li_UWCNN}\\
			
			\includegraphics[width = 0.38\linewidth]{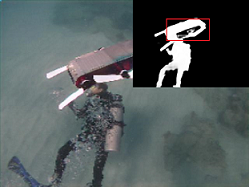} &
			\hspace{-1.6mm}
			\includegraphics[width = 0.38\linewidth]{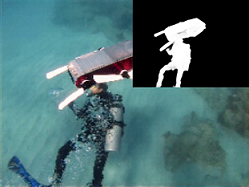}\\
			(c) FGAN~\cite{Islam_FGAN} & (d) Ours
		\end{tabular}
	\end{center}
	\vspace{-3mm}
	\caption{Saliency object detection on enhanced underwater images. WaterNet and UWCNN tend to miss detection, and FGAN tends to incomplete detection and blurry edges, while our method obtains a pleasing saliency map.}
	\label{fig:first_img}
	\vspace{-7mm}	
\end{figure}

In recent years, significant progress has been witnessed on underwater image enhancement tasks due to the use of deep CNNs~\cite{Li_UWCNN,Li_waternet,Li_Ucolor} and GANs~\cite{Islam_FGAN, Fabbri_UGAN,wang2022unsupervised}. These kinds of methods are either employed to estimate the parameters of the physical models or directly generate the enhanced images, which have shown impressive advantages in improving image contrast and alleviating color cast.
However, they fail to preserve the pleasing texture and structure information due to the limited number of datasets with imperfect hand-crafted ground truth, which hinders their application in high-level visual tasks. As shown in Fig.~\ref{fig:first_img}, the human outlines in (a) and (b) cannot be detected due to the loss of structure information, and (c) shows blurred edges due to the loss of texture details.
Thus, how to develop an effective algorithm for learning complete image structures and precise textures is very urgent for underwater enhancement tasks.

Recent researches reveal that semantic clues of the high-level vision tasks~\cite{tian2021coupled,tang2020blockmix,li2019attribute,TANG2022108792} offer guidance in low-level vision tasks, such as super-resolution~\cite{Wang_sftgan,Wang_sr_semantic}, dehazing~\cite{Ren_dehaze_semantic}, deraining~\cite{Wang_2021_icme}, and deblurring~\cite{Shen_face_semantic}. 
A common theme is that the semantic labels are exploited as global priors to guide networks to generate photo-realistic results. 
To be specific, one idea is to directly concatenate the semantic probability map with the corrupted image as the inputs of the networks~\cite{Wang_sr_semantic,Shen_face_semantic}, and the other idea is to extract features from the semantic labels to guide the decoding of features from inputs~\cite{Wang_sftgan,Ren_dehaze_semantic}.
%
Considering the extreme complexity of underwater scenes, semantic probability maps, the final products of segmentation networks, can only resolve the ambiguity in category-related object boundary but are not enough to restore accurate textures and structures required by underwater image enhancement.
Thus, we characterize the semantic-aware features of high-level visual tasks from texture and structure perspectives, and build our algorithm to exploit more informative features to facilitate underwater image enhancement.

\begin{figure*}[tbh]
	\centering
	\begin{tabular}{c}
		\includegraphics[scale=0.50]{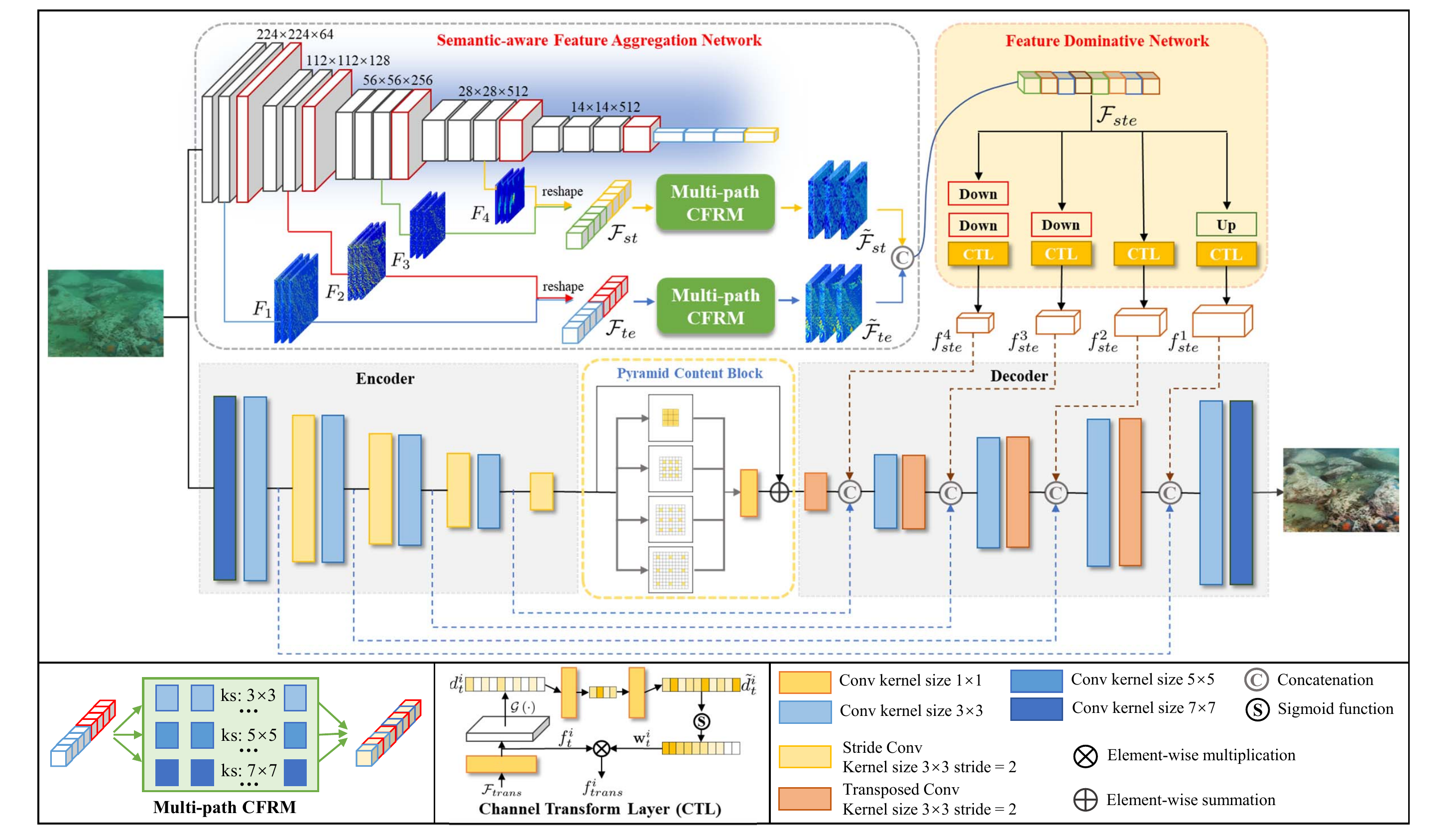} \\
	\end{tabular}
	\vspace{-2mm}
	\caption{The overview architecture of the proposed semantic-aware texture-structure feature collaboration network for underwater image enhancement.}%
	\label{fig:network}
	\vspace{-6mm}
\end{figure*}
%

%
%
Based on the above motivation, we develop a novel underwater image enhancement algorithm in cooperation with a classical pretrained semantic-aware model.  
Our algorithm framework is designed into three parts: 1) With the semantic-aware model as a powerful auxiliary, we present an efficient and compact UNet-like framework as the base network.
2) Since the generation of near-lossless image content depends on the collaboration of textures and structures, we represent features from the semantic-aware model via two independent branches according to the consensus that the shallow features imply textures while the deep features imply structures.
Besides, to refine texture and structure features in multiple scales and model correlation between features, we design and deploy a multi-path Contextual Feature Refinement Module (CFRM) on the two branches. 
%
3) Aiming at the characteristics of the non-uniform illumination and various degradation in underwater images, we present a feature dominative network to perform channel-wise modulation for the aggregated features, to adapt to different feature patterns of the base network.
%
Given that the modulated features have gone through an encoding-like process, we determine to embed them into the decoder of the base network to facilitate feature reconstruction. 
Our algorithm is a general framework, in which any semantic-aware model from high-level vision tasks (i.e., classification, semantic segmentation, and object detection) can provide important guidance for the enhancement models to restore the photo-realistic images. 
We further apply the proposed algorithm to the underwater salient object detection task. The results in Fig.~\ref{fig:first_img} reveal that our algorithm has the semantic-aware ability.
%

The main contributions of this paper are three-fold: 
\begin{itemize}	
	\item We exploit the hierarchical feature representation of the high-level semantic-aware model as an auxiliary and propose an efficient and compact underwater enhancement framework.
	
	\item We characterize semantic-aware features as textures and structures separately, and propose a multi-path contextual feature refinement module to model the correlation between features for formulating near-lossless image content. Besides, a feature dominative network is developed to perform channel-wise feature modulation to adapt to the base enhancement network.
	
	\item We conduct extensive experiments and demonstrate that the proposed algorithm performs favorably against state-of-the-art methods. The application on underwater salient detection reveals its semantic-aware ability.
	
\end{itemize}

\section{Proposed Algorithm}

\subsection{Network Architecture} 

As shown in Fig.~\ref{fig:network}, the architecture of the proposed algorithm consists of three main components, a UNet-like base network, a semantic-aware feature aggregation network, and a feature dominative network.

\textbf{UNet-like Base Network.} 
It serves as a base network for underwater image enhancement, which takes an underwater image $x$ as input and is suppose to reconstruct an enhanced image $y$ with complete structures and precise textures.
%
%
Let $E$ and $G$ denote the encoder and decoder of the base network, respectively.
The input image $x$ is first fed into the $E$ to capture hierarchical multi-scale features denoted by ${f}^{i}= E\left(x ; \Theta_{E}\right)$, where $i \in\{1,2, \ldots, {w}\}$ and ${w}$ is scale factor.
To enlarge the receptive field and maximize the information utilization, we employ a pyramid context block $\mathcal{P}\left(\cdot \right)$ at the bottom of the UNet-like network. 
Thus, the features to be decoded are noted as $\mathcal{P}\left({f}^{w}\right)$, and multi-scale reconstructed features ${g}^{i}= G\left(\mathcal{P}\left({f}^{w}\right); \Theta_{G}\right)$ are generated by the decoder.
$\Theta_{E}$ and $\Theta_{G}$ are model parameters of the $E$ and $G$.

\textbf{Semantic-aware Feature Aggregation Network.}
In this network, we exploit a high-level semantic-aware model as the feature extractor, which is a widely-used model VGG16~\cite{Simonyan_VGG16} on classification tasks.
Thanks to the training on an extra-large scale benchmark ImageNet~\cite{ImageNet}, this model embraces powerful feature representation ability and does not need to be retrained on the underwater datasets.
We directly feed $x$ into the semantic-aware model to extract hierarchical multi-scale features denoted by ${F}_{i}$, and also $i \in\{1,2, \ldots, {w}\}$.
Due to the influence of too deep semantic-aware features on low-level tasks is negligible, the value of ${w}$ is set to $4$ in all the experiments.
The core of the network is to build texture branch $\mathcal{B}_{te}$ and structure branch $\mathcal{B}_{st}$ to tackle the extracted features separately.
We represent the shallow features from the first two scales as textures and represent the deep features from the last two scales as structures, and reorganize them via reshaping operation and concatenation. 
For the reorganized texture features $\mathcal{F}_{te}$ and structure features $\mathcal{F}_{st}$, we present a multi-path Contextual Feature Refinement Module (CFRM) to implement two refinement processes $ \mathcal{F}_{te} \rightarrow \tilde{\mathcal{F}_{te}}$ and $\mathcal{F}_{st} \rightarrow \tilde{\mathcal{F}_{st}}$.
%
%
As shown in Fig.~\ref{fig:network}, there are three parallel convolution paths with progressively increased kernel sizes in the CFRM.
The refined texture and structure features by each path are denoted as ${\phi}_{te}^{k}\left(\mathcal{F}_{te} \right)$ and ${\phi}_{st}^{k}\left(\mathcal{F}_{st} \right)$, where $k \in K$ and $K=[3,5,7]$.
%
Note that both ${\phi}_{te}^{k}\left(\cdot \right)$ and ${\phi}_{st}^{k}\left(\cdot \right)$ denote the convolution paths with kernel size of $k\times k$ in $\mathcal{B}_{te}$ and $\mathcal{B}_{st}$, respectively.
Thus, the textures and structures refined by the CFRMs are represented as 
\begin{equation}
	\vspace{-1.0mm}
	\begin{split}
		\tilde{\mathcal{F}}_{te} =\mathcal{C}\left({\phi}_{te}^{k}\left(\mathcal{F}_{te} \right),k \in K \right),\\	
		\tilde{\mathcal{F}}_{st} =\mathcal{C}\left({\phi}_{st}^{k}\left(\mathcal{F}_{st} \right),k \in K\right).
	\vspace{-1.0mm}
	\end{split}
\end{equation}
%
Subsequently, we aggregate the texture and structure features by 
\begin{equation}
	\vspace{-1.0mm}
	\mathcal{F}_{ste} = \mathcal{C}\left(\tilde{\mathcal{F}}_{te}, \tilde{\mathcal{F}}_{st} \right),
	\vspace{-1.0mm}
\end{equation}
where $\mathcal{C}$ denotes the concatenation and a followed $1\times1$ convolution layer to reduce the feature channels.

\textbf{Feature Dominative Network.}
Due to the characteristics of non-uniform illumination and various degradation in underwater images, it is not applicable to directly embed the aggregated features into the base network.
Hence, we present a feature dominative network to learn image-specific and region-specific features through performing channel-wise feature modulation.
To match features with different scales of the base network, the aggregated features $\mathcal{F}_{ste}$ are first reshaped into those with specific sizes, which are then fed to the Channel Transformation Layer (CTL) to obtain applicable feature embeddings matching the base network.
As shown in Fig.~\ref{fig:network}, a convolution layer of $1\times 1$ is fist used to conduct a transformation $\mathcal{F}_{ste}\rightarrow f_{t}^{i}$, also $i \in\{1,2, \ldots, {w}\}$. And then an adaptive weight vector $\mathbf{w}_{t}^{i}$ is obtained by global average pooling, down-upscaling operations, and sigmoid function. The final modulated features $f_{ste}^{i}$ are obtained by ${f}_{t}^{i} \otimes \mathbf{w}_{t}^{i}$, where $\otimes$ represents the element-wise multiplication.

Finally, we embed modulated features into the decoder of the UNet-like base network to generate an enhanced underwater image by
\begin{equation}
	y = G_{i\rightarrow w}\left(f^{i\rightarrow w}, f_{ste}^{i\rightarrow w}, g^{i\rightarrow w} \right),
\end{equation}
where ${i\rightarrow w}$ is the scale range of the features. $G_{i\rightarrow w}$ denotes the progressive feature reconstruction by the decoder under the scale range.

\subsection{Loss Functions}

Considering that human visual perception often pays more attention to image details and textures, we use multi-scale structure similarity (MS-SSIM~\cite{Zhao_ms_ssim}) loss function $\mathcal{L}_{ssim}^{sm}$ to optimize our network, thus generate more realistic image content. However, doing so leads to image contrast change and color distortion. In view of this, we use a widely-used pixel-wise loss function $\mathcal{L}_{1}$ to ensure the insensitivity of the network to image contrast and color, so as to achieve a good trade-off between content restoration and picture fidelity. Therefore, the loss function is defined as:
\begin{equation}
	\label{eq: l1-loss-function}
	\mathcal{L} = \lambda \cdot \mathcal{L}_{ssim}^{ms} + \left(1 - \lambda \right) \cdot \mathcal{L}_{1},
\end{equation}
where, $\lambda$ is a balance parameter. It is set to $0.8$ in our work.

\section{EXPERIMENTAL RESULTS}

\subsection{Dataset and Implementation Details}
\label{sec:dataset}
\textbf{Datasets.}
We evaluate the proposed algorithm using two labeled underwater datasets UIEB~\cite{Li_waternet} and EUVP~\cite{Islam_FGAN}, and an unlabeled real-world underwater dataset RUIE~\cite{Liu_RUIE}, respectively. Moreover, we also utilize a new challenging underwater salient object detection dataset USOD~\cite{islam_USOD} to verify the semantic-aware ability of our algorithm on the high-level vision tasks. Table~\ref{tab:dataset} provides a detailed description of these datasets.
\begin{table}[t]
	\vspace{-2mm}
	\centering
	\caption{\label{tab:dataset}Description of the benchmark underwater datasets.}
	\vspace{-2mm}
	\setlength{\tabcolsep}{2.6mm}{
		\begin{tabular}{lccc}
			\toprule
			\specialrule{0em}{1pt}{1pt}
			\textbf{Datasets}  & \textbf{Training (\#)}  & \textbf{Testing (\#)} & \textbf{Paired/Unpaired} \\
			\hline
			\specialrule{0em}{1pt}{1pt}
			UIEB~\cite{Li_waternet}  & $712$  & $238$  & Paired\\
			\specialrule{0em}{1pt}{1pt}
			EUVP~\cite{Islam_FGAN}  & $7,200$  & $4,284$  & Paired\\
			\specialrule{0em}{1pt}{1pt}
			RUIE~\cite{Liu_RUIE}      & $0$ & $300$ & Unpaired\\
			\specialrule{0em}{1pt}{1pt}
			USOD~\cite{islam_USOD}     & $0$ & $300$ & Paired\\
			\bottomrule
	\end{tabular}}
	\vspace{-6mm}
\end{table}

\textbf{Training Settings.}
We randomly crop $8$ image patches of size $224\times 224$ to form a batch. 
The Adam optimizer~\cite{Kingma_Adam} ($\beta _{1}=0.9$, and $\beta _{2}=0.999$) is used to optimize our model. The initial learning rate is $5\times10^{^{-4}}$, and decreasing to $0.2$ times every $8,000$ iterations during training. 
The whole training phase goes through $100,000$ iterations. We implement the proposed network using the PyTorch framework with an NVIDIA 1080Ti GPU.

\textbf{Evaluation Metrics.}
For the sake of the comprehensive and fair assessment, we employ four metrics involving reference and non-reference approaches. For the UIEB~\cite{Li_waternet} and EUVP~\cite{Islam_FGAN} dataset with reference images, we mainly adopt two widely-used metrics (i.e., PSNR (dB) and SSIM) for evaluation. For the RUIE~\cite{Liu_RUIE} dataset without reference images, we mainly use the other two non-reference evaluation metrics (i.e., Underwater Image Quality Measurement UIQM~\cite{Karen_uiqm} and Natural Image Quality Evaluator NIQE~\cite{Anish_niqe}).

\subsection{Comparison with State-of-the-art Methods}

We conduct extensive experiments to quantitatively and qualitatively evaluate our algorithm against several state-of-the-art methods including conventional methods~\cite{Fu_TSA,Li_OCM,Ancuti_EUIVF}, physical model-based methods~\cite{Uplavikar_AIO,Drews_UDCP}, and data-driven deep learning-based methods~\cite{Li_waternet, Li_UWCNN, Islam_FGAN, Fabbri_UGAN,Li_Ucolor}.


\begin{table*}[tbh]
	\footnotesize
	\caption{Quantitative PSNR and SSIM values of different methods on real-world benchmark datasets (UIEB and EUVP). The value with \textcolor{red}{red} bold font indicates ranking the first place in this row while the value with \textcolor{blue}{blue} font is the second place.}
	\label{tab:psnr-ssim}
	\vspace{-2mm}
	\centering
	\setlength{\tabcolsep}{1.01mm}{
		\renewcommand{\arraystretch}{1.1}
		\begin{tabular}{l|c|cccccccccc|c}
			\toprule
			Dataset & Metric & EUIVF~\cite{Ancuti_EUIVF} & OCM~\cite{Li_OCM} & UDCP~\cite{Drews_UDCP} & TSA~\cite{Fu_TSA} & UGAN~\cite{Fabbri_UGAN} & WaterNet~\cite{Li_waternet} & AIO~\cite{Uplavikar_AIO} & FGAN~\cite{Islam_FGAN} & UWCNN~\cite{Li_UWCNN} & Ucolor~\cite{Li_Ucolor} & Ours\\
			\hline
			\multirow{2}{*}{UIEB} & PSNR & $\textcolor{blue}{\textbf{21.93}}$ & $16.19$ & $11.73$ & $14.32$ & $17.73$ &  $19.65$ & $12.69$ & $18.16$ & $13.35$ & $20.62$ & $\textcolor{red}{\textbf{22.45}}$\\
			& SSIM & $0.823$ & $0.759$ &  $0.509$ & $0.763$ & $0.765$ & $0.824$ & $0.466$ & $0.597$ & $0.773$ & $\textcolor{red}{\textbf{0.921}}$ & $\textcolor{blue}{\textbf{0.902}}$\\
			\hline
			\multirow{2}{*}{EUVP} & PSNR & $17.06$ & $15.62$ & $14.53$ & $13.21$ & $19.31$ & $18.68$ & $16.25$ & $\textcolor{blue}{\textbf{19.49}}$ & $18.37$ & $--$ & $\textcolor{red}{\textbf{23.23}}$\\
			& SSIM  & $0.894$ & $0.843$ & $0.888$ & $0.672$ & $0.890$ & $0.952$ & $0.881$ & $\textcolor{blue}{\textbf{0.963}}$ & $0.948$ & $--$ & $\textcolor{red}{\textbf{0.987}}$ \\
			\bottomrule
	\end{tabular}}
	\vspace{-6mm}
\end{table*}

\textbf{Quantitative Evaluation.}
We report the quantitative results on UIEB, EUVP, and RUIE benchmarks in Table~\ref{tab:psnr-ssim} and Table~\ref{tab:niqe-uiqm}. It can be seen that our algorithm numerically outperforms most existing methods by large margins and ranks first or second in the four evaluation metrics. 
Especially, compared with Ucolor~\cite{Li_Ucolor}, a recent research, our algorithm gains $\textbf{1.83}$\textbf{dB} and $\textbf{0.982}$ improvements in PSNR and UIQM on the UIEB dataset. Due to the fact that transmission maps are required by Ucolor and the code~\cite{Peng_GDCP} for estimating them has not been released, we cannot make comparisons with it on EUVP and RUIE datasets. 
In addition, our algorithm performs better than current prevalent data-driven methods in all four metrics, such as WaterNet~\cite{Li_waternet} and UWCNN~\cite{Li_UWCNN}.

\begin{figure*}[tbh]
	\begin{center}
		\begin{tabular}{cccccc}
			\includegraphics[width = 0.152\linewidth]{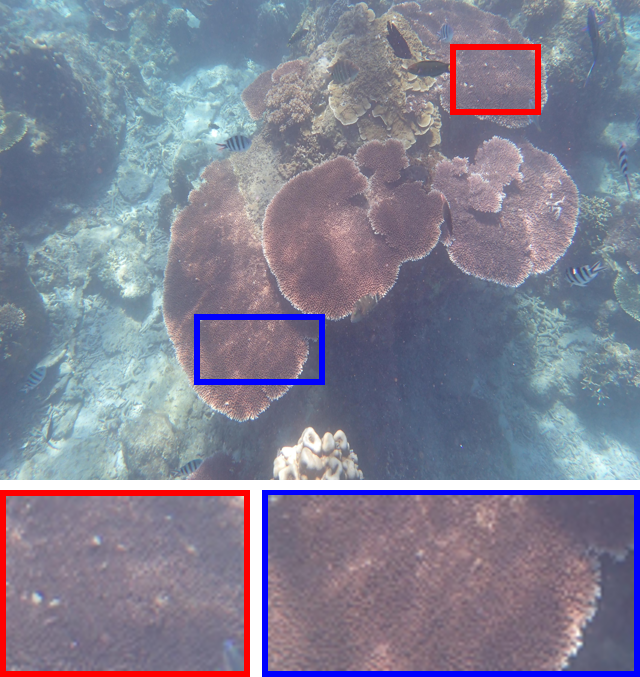}  &
			\hspace{-0.46cm}
			\includegraphics[width = 0.152\linewidth]{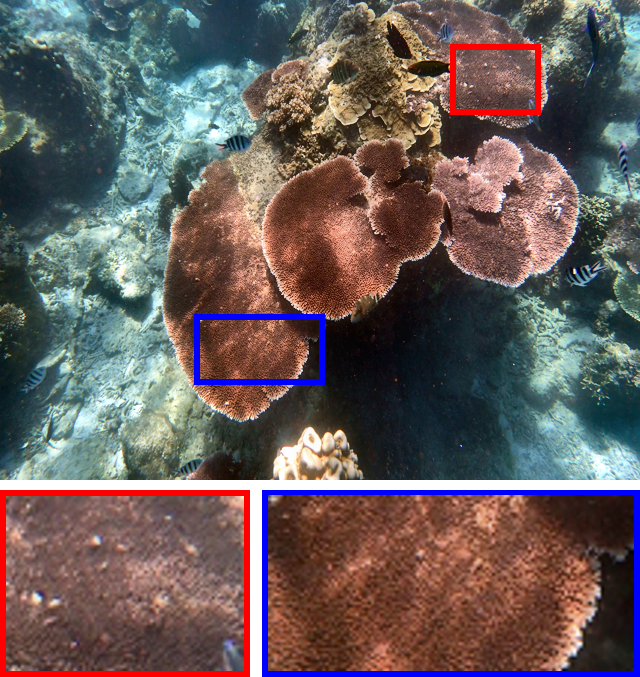}  & \hspace{-0.46cm}
			\includegraphics[width = 0.152\linewidth]{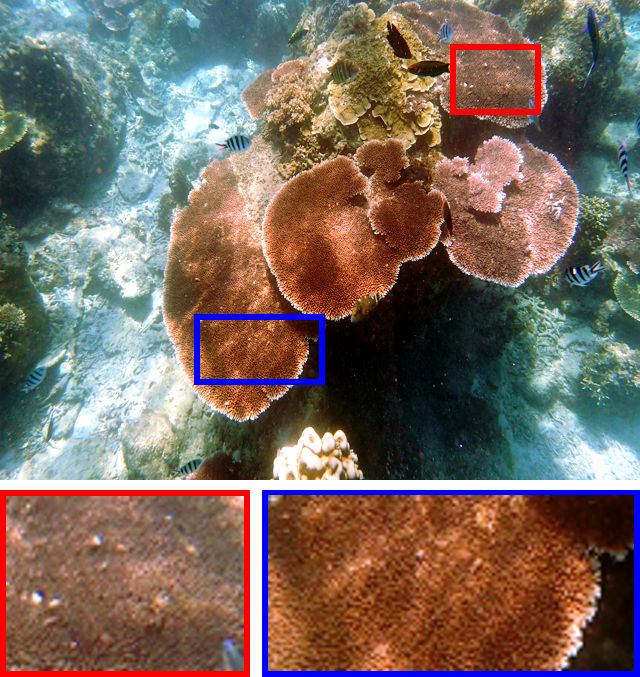}    & \hspace{-0.46cm}
			\includegraphics[width = 0.152\linewidth]{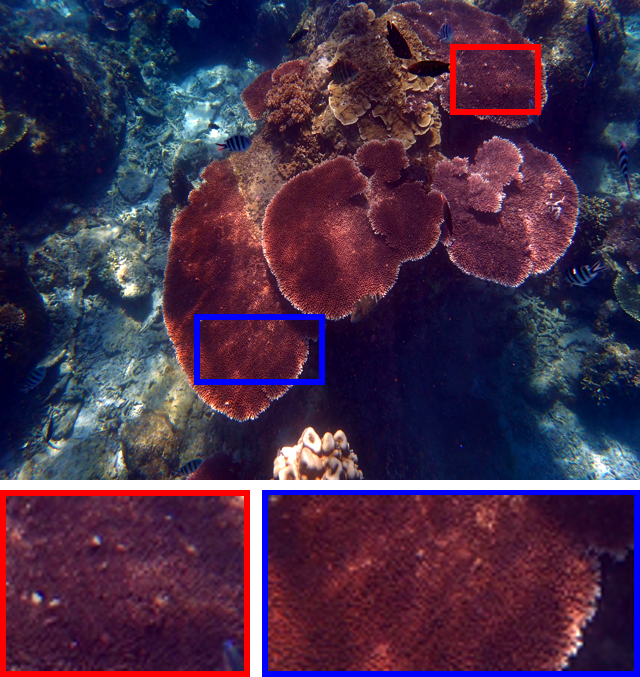}   & \hspace{-0.46cm}
			\includegraphics[width = 0.152\linewidth]{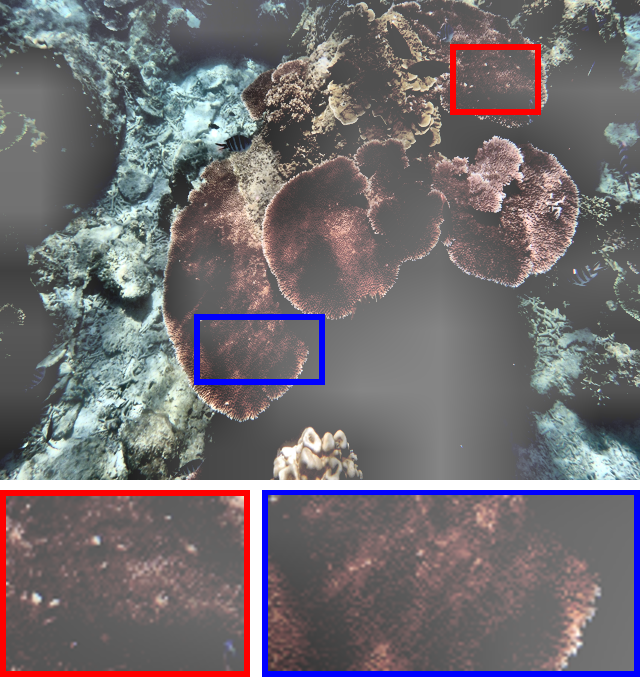}    &
			\hspace{-0.46cm}
			\includegraphics[width = 0.152\linewidth]{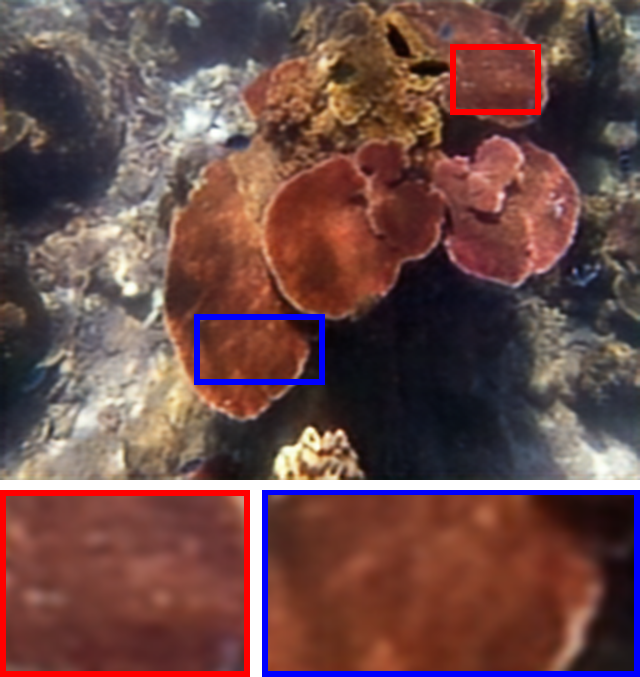} \\		
			
			\small(a) Input
			& \hspace{-0.36cm} \small(b) EUIVF 
			& \hspace{-0.36cm} \small(c) OCM
			& \hspace{-0.36cm} \small(d) UDCP
			& \hspace{-0.36cm} \small(e) TSA
			& \hspace{-0.36cm} \small(f) UGAN  \\
			
			\includegraphics[width = 0.152\linewidth]{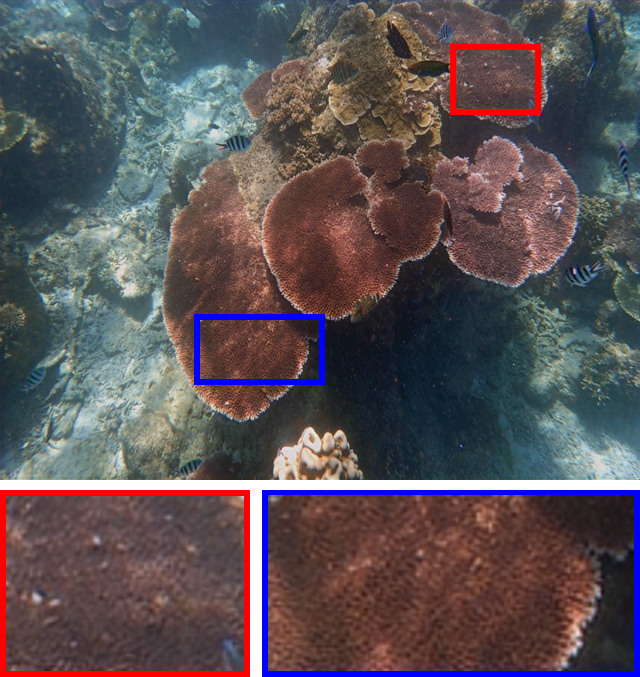}  &
			\hspace{-0.46cm}
			\includegraphics[width = 0.152\linewidth]{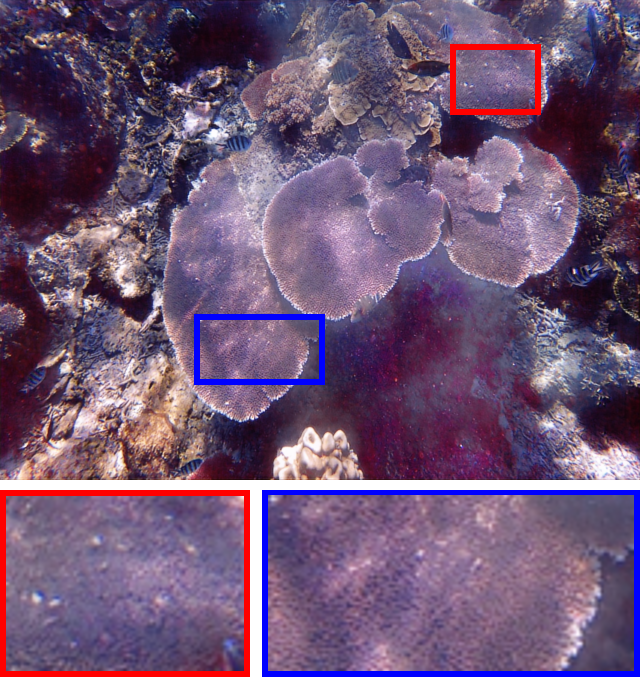}    & \hspace{-0.46cm}
			\includegraphics[width = 0.152\linewidth]{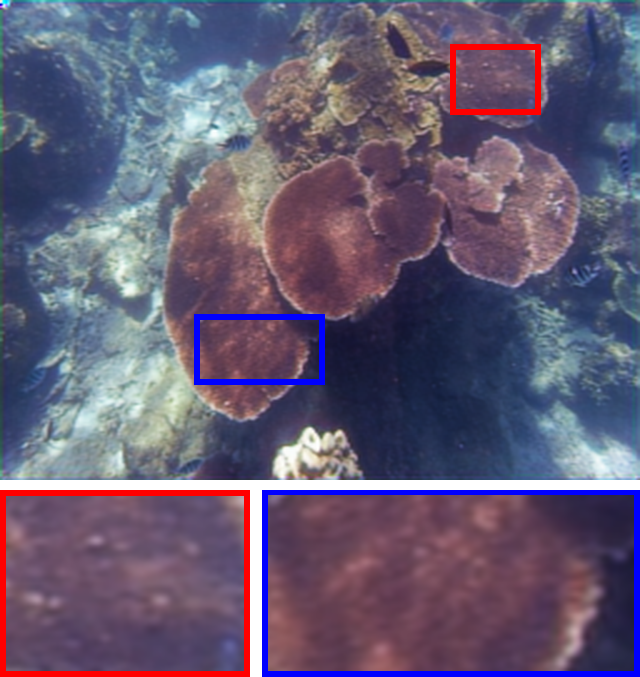}  & \hspace{-0.46cm}
			\includegraphics[width = 0.152\linewidth]{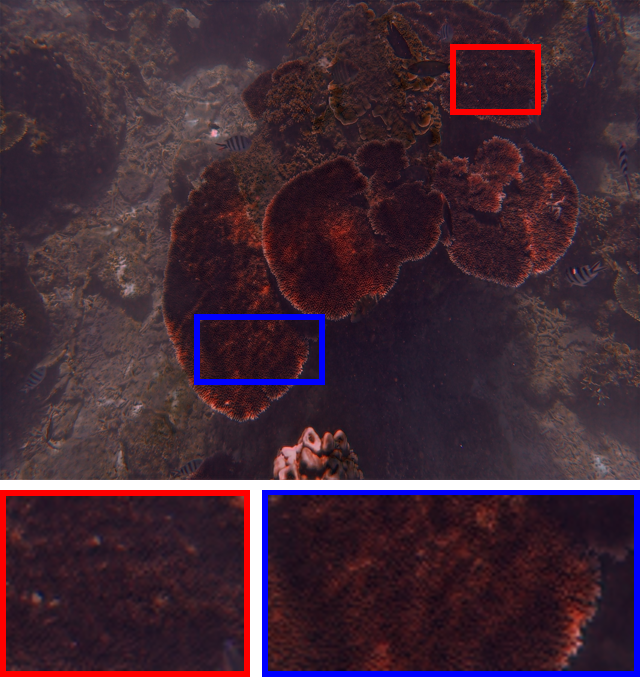}   & \hspace{-0.46cm}
			\includegraphics[width = 0.152\linewidth]{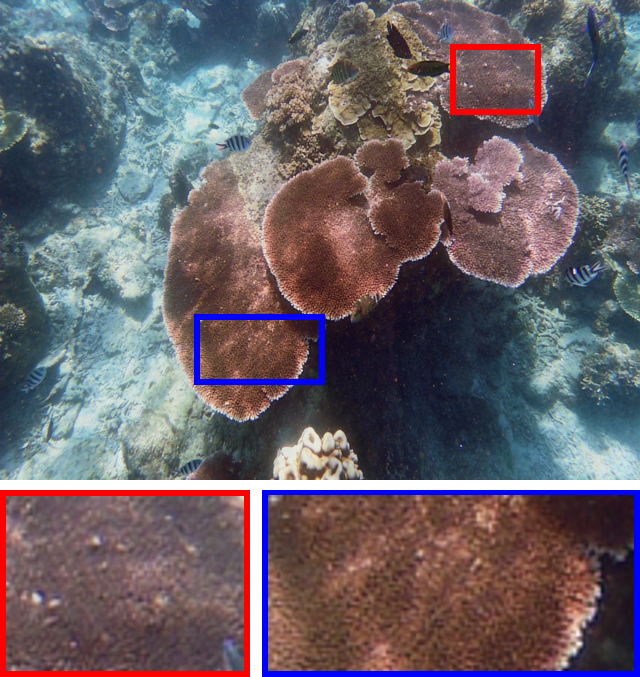}    &
			\hspace{-0.46cm}
			\includegraphics[width = 0.152\linewidth]{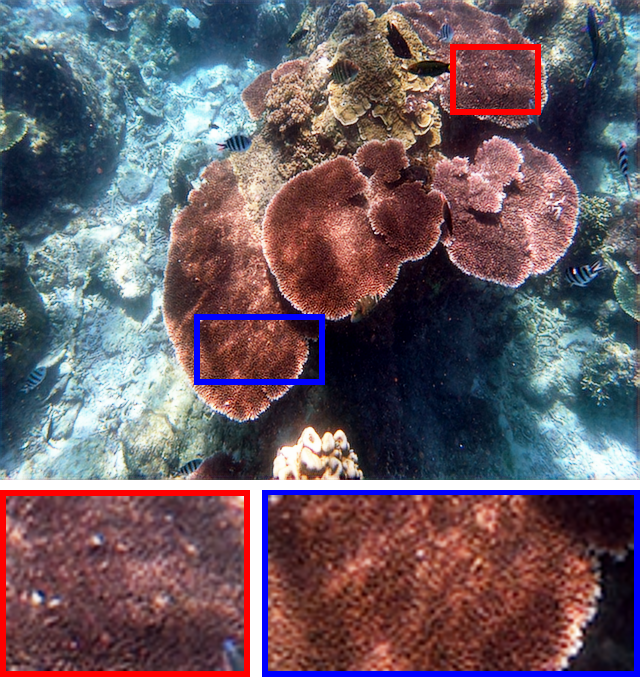} \\	
			
			\small(g) WaterNet
			& \hspace{-0.36cm} \small(h) AIO
			& \hspace{-0.36cm} \small(i) FGAN
			& \hspace{-0.36cm} \small(j) UWCNN
			& \hspace{-0.36cm} \small(k) Ucolor
			& \hspace{-0.36cm} \small(l) Ours \\			
			
		\end{tabular}
	\end{center}
	\vspace{-3mm}
	\caption{\label{fig:visual-uieb}Qualitative comparisons on the UIEB dataset. The enhanced result by our algorithm has more pleasing contrast and more precise textures.}
\end{figure*}

\begin{figure*}[h]
	\begin{center}
		\vspace{-2mm}
		\begin{tabular}{cccccccccc}
			\includegraphics[width = 0.09\linewidth]{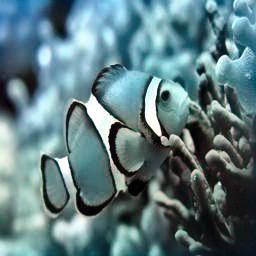}    & \hspace{-0.46cm}
			\includegraphics[width = 0.09\linewidth]{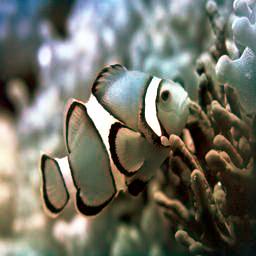}      & \hspace{-0.46cm}
			\includegraphics[width = 0.09\linewidth]{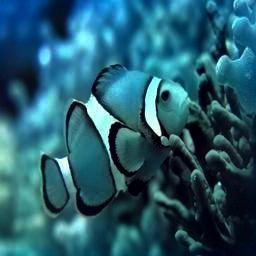}     & \hspace{-0.46cm}
			\includegraphics[width = 0.09\linewidth]{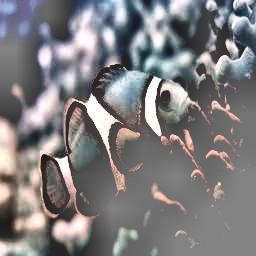}      &
			\hspace{-0.46cm}
			\includegraphics[width = 0.09\linewidth]{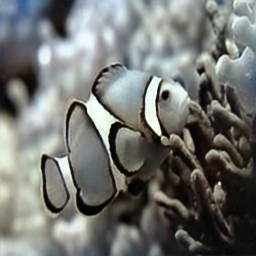}     &
			\hspace{-0.46cm}
			\includegraphics[width = 0.09\linewidth]{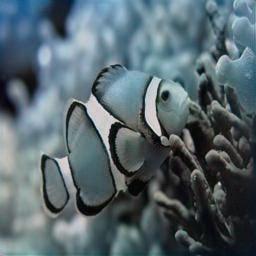} &
			\hspace{-0.46cm}
			\includegraphics[width = 0.09\linewidth]{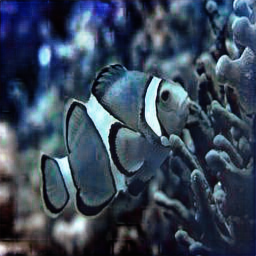}      & \hspace{-0.46cm}
			\includegraphics[width = 0.09\linewidth]{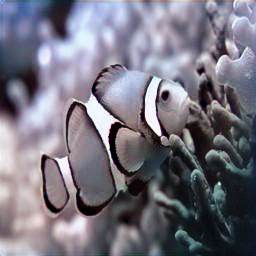}     & \hspace{-0.46cm}
			\includegraphics[width = 0.09\linewidth]{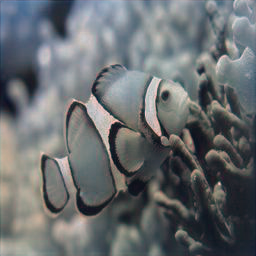}    & \hspace{-0.46cm}
			\includegraphics[width = 0.09\linewidth]{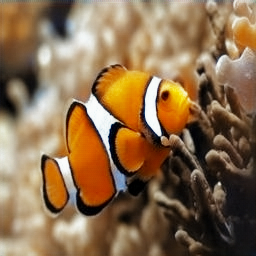} \\
			
			\includegraphics[width = 0.09\linewidth]{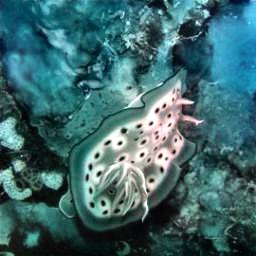}    & \hspace{-0.46cm}
			\includegraphics[width = 0.09\linewidth]{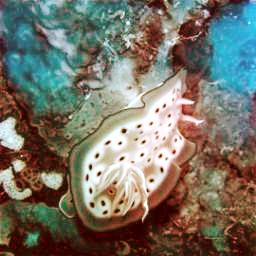}      & \hspace{-0.46cm}
			\includegraphics[width = 0.09\linewidth]{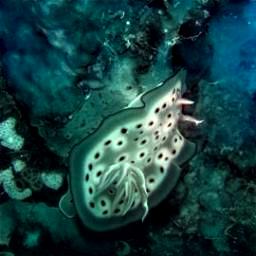}     & \hspace{-0.46cm}
			\includegraphics[width = 0.09\linewidth]{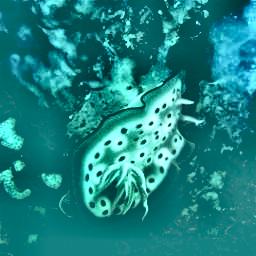}      &
			\hspace{-0.46cm}
			\includegraphics[width = 0.09\linewidth]{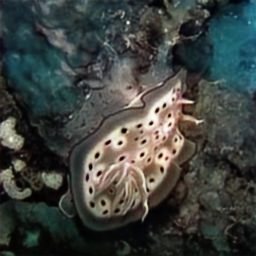}     &
			\hspace{-0.46cm}
			\includegraphics[width = 0.09\linewidth]{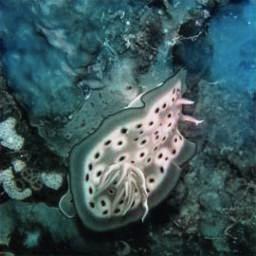} &
			\hspace{-0.46cm}
			\includegraphics[width = 0.09\linewidth]{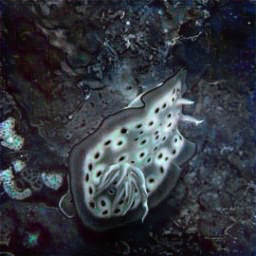}      & \hspace{-0.46cm}
			\includegraphics[width = 0.09\linewidth]{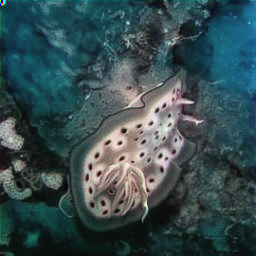}     & \hspace{-0.46cm}
			\includegraphics[width = 0.09\linewidth]{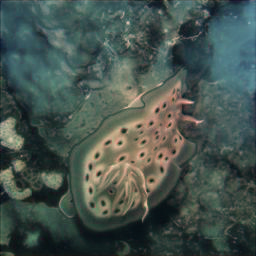}    & \hspace{-0.46cm}
			\includegraphics[width = 0.09\linewidth]{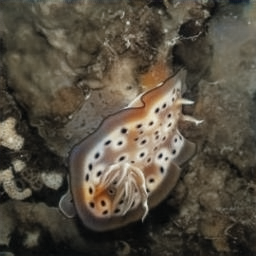} \\

			\small EUIVF
			& \hspace{-0.36cm} \small OCM
			& \hspace{-0.36cm} \small UDCP
			& \hspace{-0.36cm} \small TSA
			& \hspace{-0.36cm} \small UGAN
			& \hspace{-0.36cm} \small WaterNet
			& \hspace{-0.36cm} \small AIO
			& \hspace{-0.36cm} \small FGAN
			& \hspace{-0.36cm} \small UWCNN
			& \hspace{-0.36cm} \small Ours
			\\
			
		\end{tabular}
	\end{center}
	\vspace{-4mm}
	\caption{\label{fig:visual-euvp}Qualitative comparisons on the EUVP dataset. The enhanced results by our algorithm have better textures and colors.}
	\vspace{-1mm}
\end{figure*}

\begin{figure*}[h]\footnotesize
	\begin{center}
		\vspace{-2mm}
		\begin{tabular}{cccccccc}
			\multicolumn{3}{c}{\multirow{5}*[51.0pt]{\includegraphics[width=0.28\linewidth, height = 0.250\linewidth]{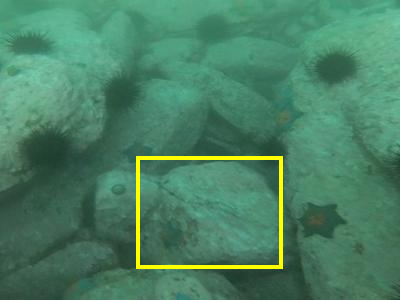}}}&\hspace{-4mm}
			\includegraphics[width=0.128\linewidth, height = 0.120\linewidth]{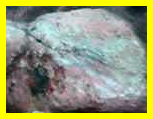} &\hspace{-4mm}
			\includegraphics[width=0.128\linewidth, height = 0.120\linewidth]{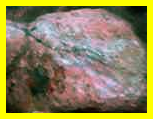} &\hspace{-4mm}
			\includegraphics[width=0.128\linewidth, height = 0.12\linewidth]{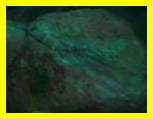} &\hspace{-4mm}
			\includegraphics[width=0.128\linewidth, height = 0.12\linewidth]{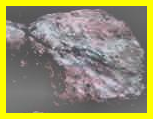}  &\hspace{-4mm}
			\includegraphics[width=0.128\linewidth, height = 0.12\linewidth]{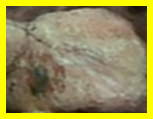}
			\\
			\multicolumn{3}{c}{~} &\hspace{-4mm}  \small(b) EUIVF &\hspace{-4mm}  \small(c) OCM &\hspace{-4mm}  \small(d) UDCP &\hspace{-4mm}  \small(e) TSA \hspace{-4mm} & \small(f) UGAN \\
			\multicolumn{3}{c}{~} & \hspace{-4mm}
			\includegraphics[width=0.128\linewidth, height = 0.120\linewidth]{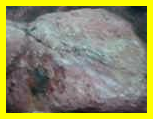} & \hspace{-4mm}
			\includegraphics[width=0.128\linewidth, height = 0.120\linewidth]{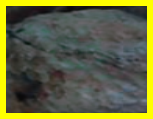} & \hspace{-4mm}
			\includegraphics[width=0.128\linewidth, height = 0.120\linewidth]{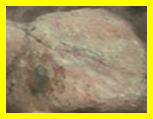} & \hspace{-4mm}
			\includegraphics[width=0.128\linewidth, height = 0.120\linewidth]{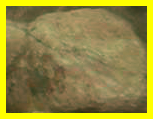} & \hspace{-4mm}
			\includegraphics[width=0.128\linewidth, height = 0.120\linewidth]{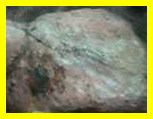} \\
			
			\multicolumn{3}{c}{\hspace{-4mm} \small(a) Input} &  \hspace{-4mm} \small(g) WaterNet &\hspace{-4mm}  \small(h) AIO &\hspace{-4mm} \small (i) FGAN & \hspace{-4mm} \small(j) UWCNN & \hspace{-4mm} \small(k) Ours\\
		\end{tabular}
	\end{center}
	\vspace{-3mm}
	\caption{Visualization of the underwater enhancement results by different methods on the RUIE dataset.}%
	\label{fig: visual-ruie}
	\vspace{-2mm}
\end{figure*}

\textbf{Qualitative Evaluation.}
We first show the qualitative evaluations on the UIEB benchmark in Fig.~\ref{fig:visual-uieb}. By observing the local enlarged areas, we note that some methods such as TSA~\cite{Fu_TSA}, AIO~\cite{Uplavikar_AIO}, WaterNet~\cite{Li_waternet}, and FGAN~\cite{Islam_FGAN}, can not effectively alleviate the underwater haze-effect, while the UDCP~\cite{Drews_UDCP} and UWCNN~\cite{Li_UWCNN} cause severe contrast reduction and color cast. More seriously, almost all the comparison methods fail to restore the complete structures and precise textures. In contrast, the underwater image enhanced by our algorithm has a sharper structure and richer texture and achieves a balance of contrast and color cast simultaneously.
Then, we also show qualitative results on the EUVP benchmark in Fig.~\ref{fig:visual-euvp}.
It can be observed that our algorithm performs well in dealing with structure characteristics such as contrast and color, and the precise textures are also clearly displayed incidentally.
%
Moreover, we also show some visualization examples on the RUIE benchmark in Fig.~\ref{fig: visual-ruie}. The greenish color cast weakens structure information and hides texture details of the underwater scenes as shown in (a). According to (b)-(k), we can observe that these visual results are either under-enhanced or introduce the reddish and brownish color cast, while our algorithm shows relatively more realistic textures.

\begin{figure}[t]
	\vspace{-3mm}
	\centering
	\begin{tabular}{c}
		\includegraphics[width=0.75\linewidth]{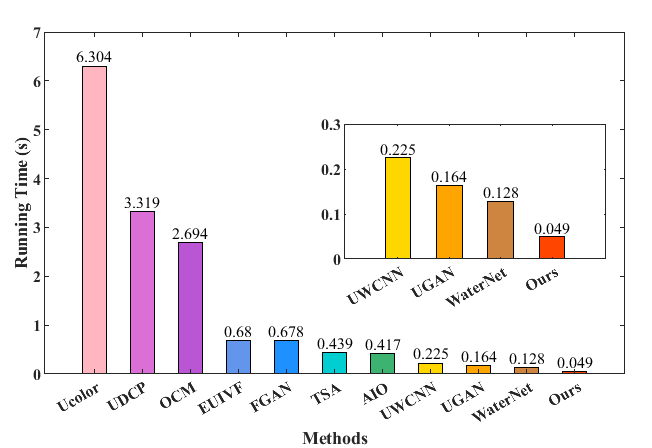} \\
	\end{tabular}
	\vspace{-3mm}
	\caption{Running time comparisons against state-of-the-art methods on a color image of size $648 \times 480$.}%
	\label{fig:run-time}
	\vspace{-7mm}
\end{figure}

\textbf{Time Complexity Evaluation.}
For the efficiency of our algorithm, we compare time complexity against the state-of-the-art models on a single 1080Ti GPU. As shown in Fig.~\ref{fig:run-time}, our model performs faster, especially compared to the latest underwater enhancement method Ucolor~\cite{Li_Ucolor}. Combined with ahead experimental results, it can be illustrated that the proposed algorithm can obtain desirable underwater enhancement results at a low computational cost.

\subsection{Application for other High-level Tasks}

\begin{figure*}[h]
	\begin{center}
		\begin{tabular}{cccccccccccc}	
			
			\includegraphics[width = 0.082\linewidth]{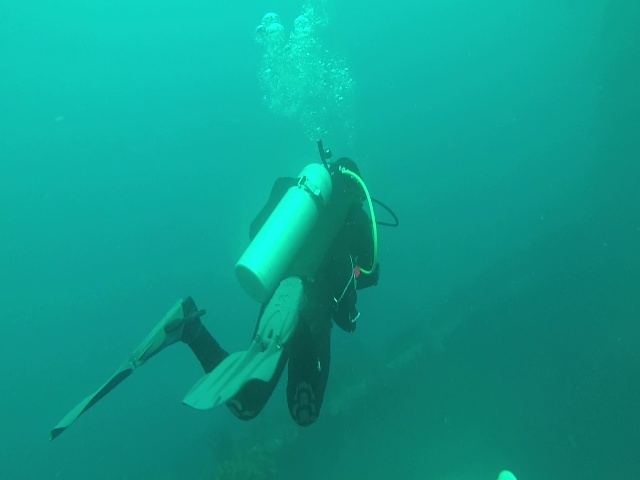}        & \hspace{-0.46cm}
			\includegraphics[width = 0.082\linewidth]{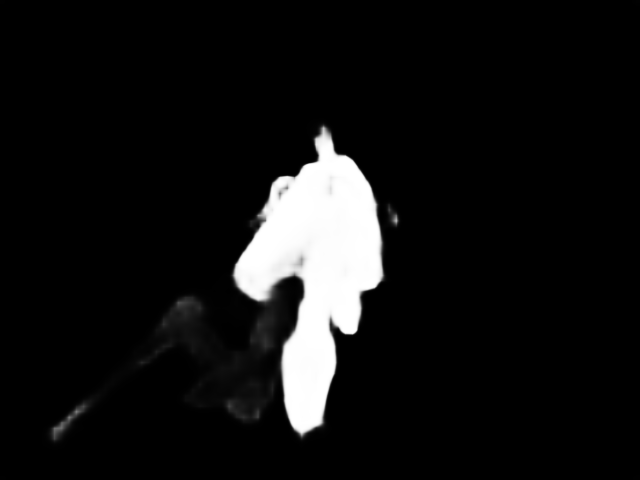}  & \hspace{-0.46cm}
			\includegraphics[width = 0.082\linewidth]{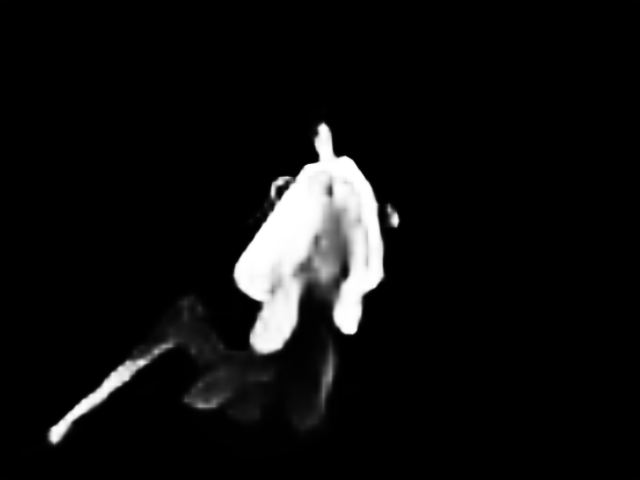}   & \hspace{-0.46cm}
			\includegraphics[width = 0.082\linewidth]{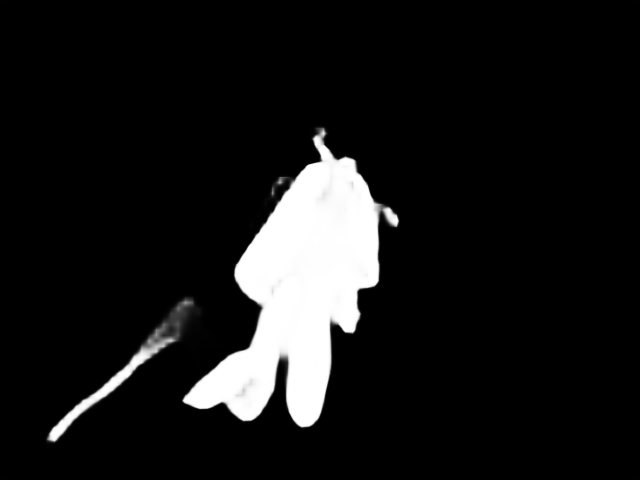} & \hspace{-0.46cm}
			\includegraphics[width = 0.08\linewidth]{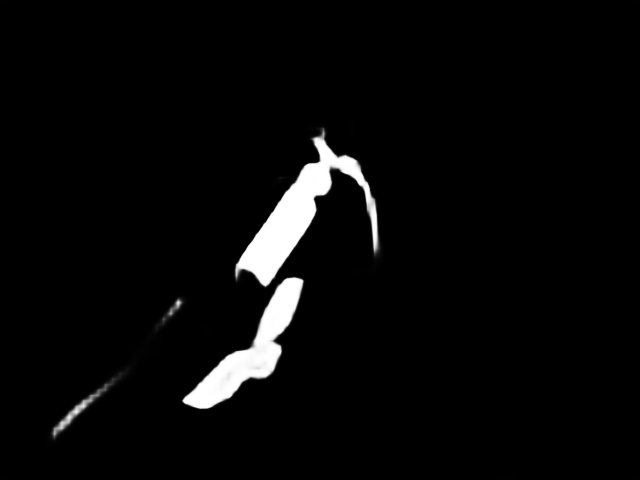} &
			\hspace{-0.46cm}
			\includegraphics[width = 0.082\linewidth]{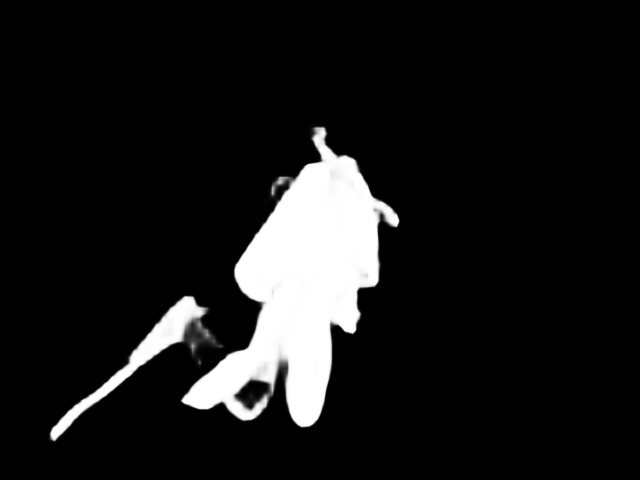} &
			\hspace{-0.46cm}
			\includegraphics[width = 0.082\linewidth]{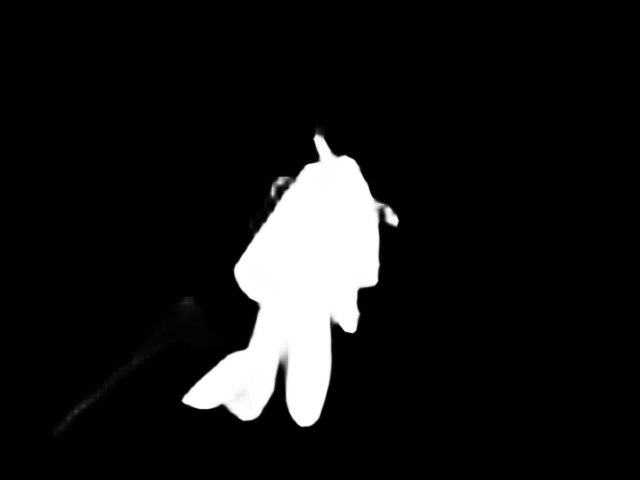} &
			\hspace{-0.46cm}
			\includegraphics[width = 0.082\linewidth]{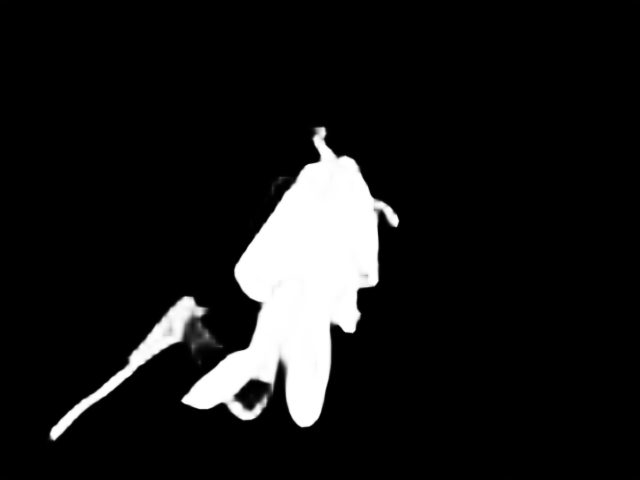} &
			\hspace{-0.46cm}
			\includegraphics[width = 0.082\linewidth]{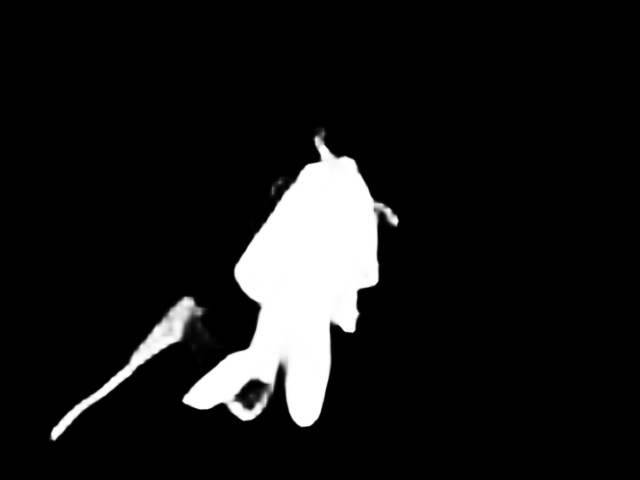} &
			\hspace{-0.46cm}
			\includegraphics[width = 0.082\linewidth]{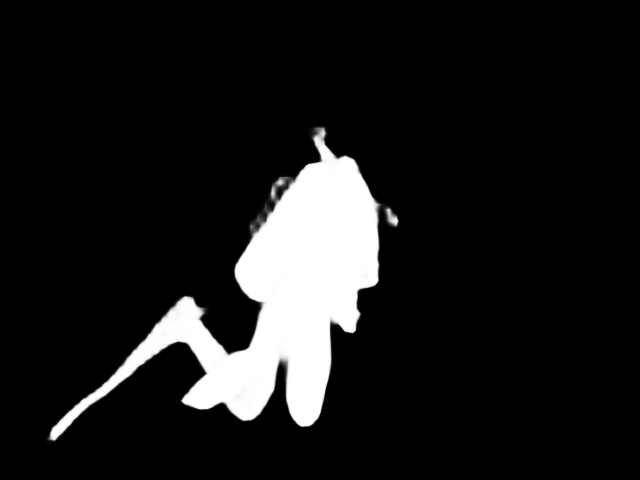} &
			\hspace{-0.46cm}
			\includegraphics[width = 0.082\linewidth]{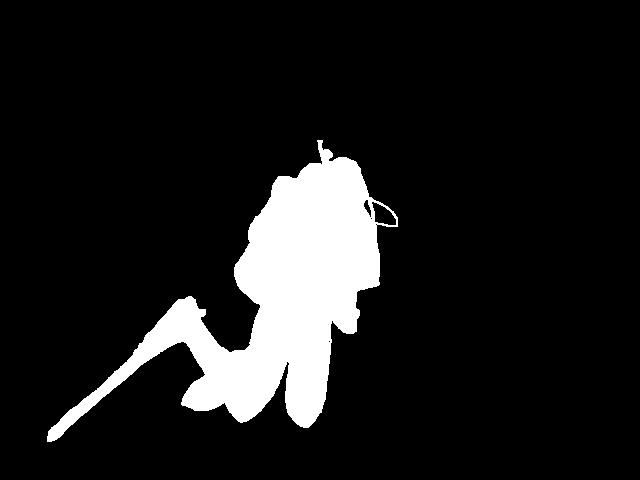} \\	
			
			\includegraphics[width = 0.082\linewidth]{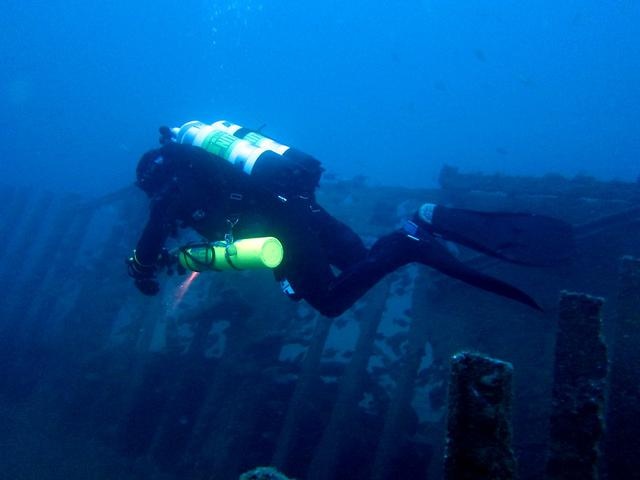}        & \hspace{-0.46cm}
			\includegraphics[width = 0.082\linewidth]{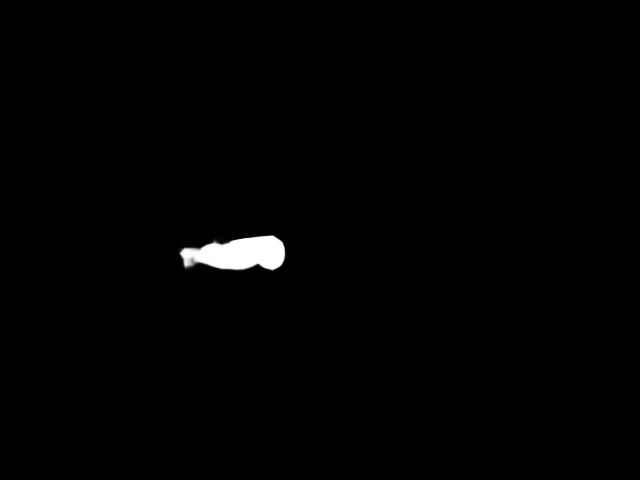}  & \hspace{-0.46cm}
			\includegraphics[width = 0.082\linewidth]{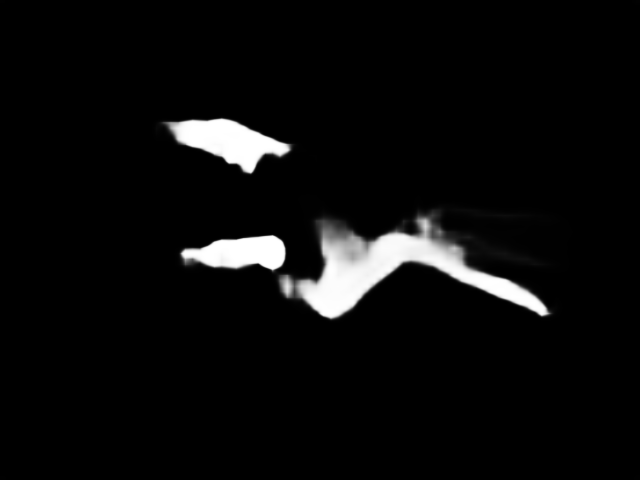}   & \hspace{-0.46cm}
			\includegraphics[width = 0.082\linewidth]{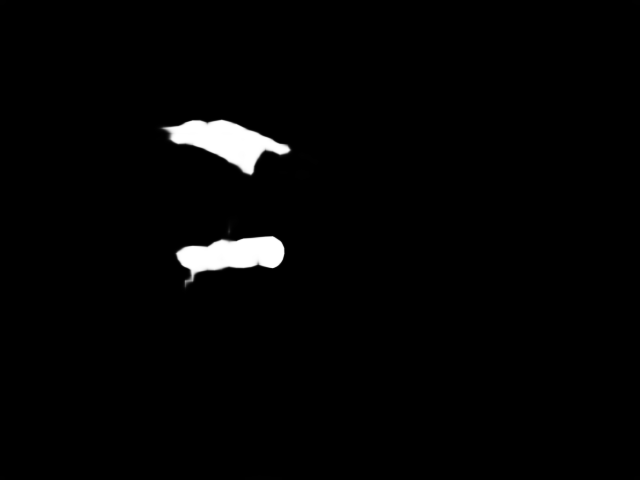} & \hspace{-0.46cm}
			\includegraphics[width = 0.082\linewidth]{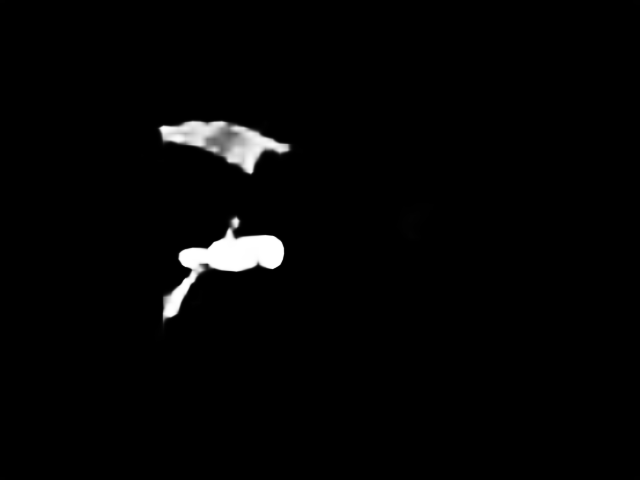} &
			\hspace{-0.46cm}
			\includegraphics[width = 0.082\linewidth]{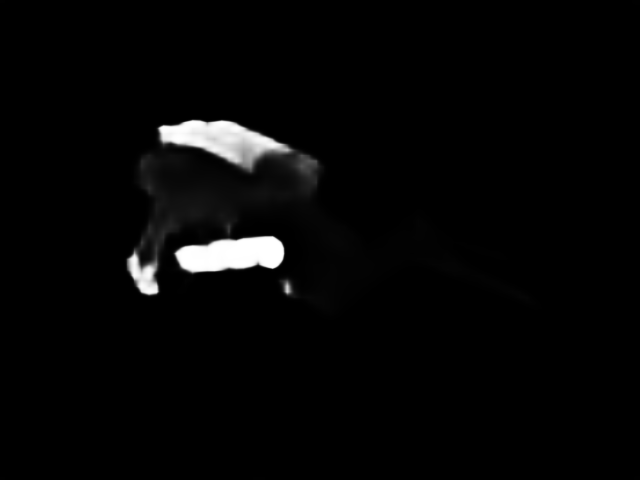} &
			\hspace{-0.46cm}
			\includegraphics[width = 0.082\linewidth]{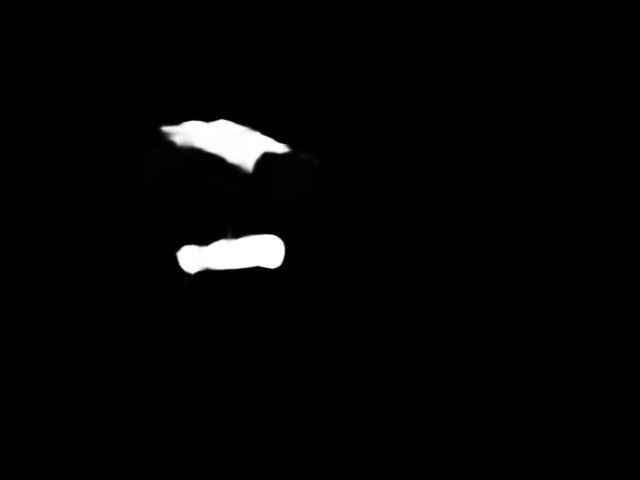} &
			\hspace{-0.46cm}
			\includegraphics[width = 0.082\linewidth]{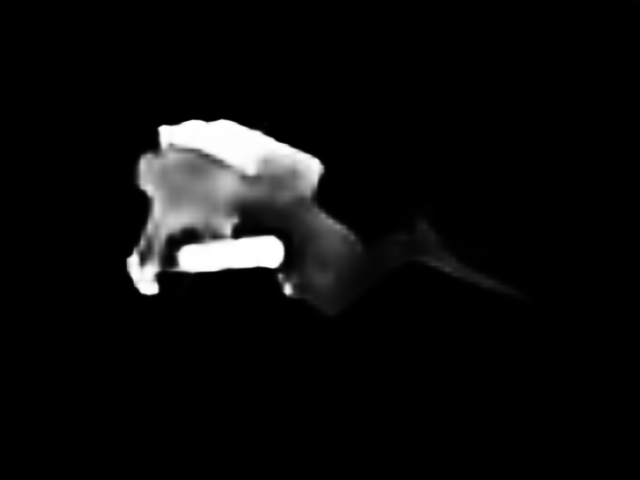} &
			\hspace{-0.46cm}
			\includegraphics[width = 0.082\linewidth]{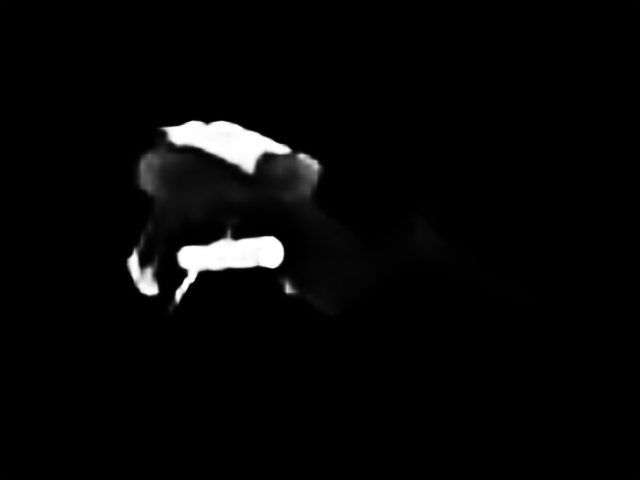} &
			\hspace{-0.46cm}
			\includegraphics[width = 0.082\linewidth]{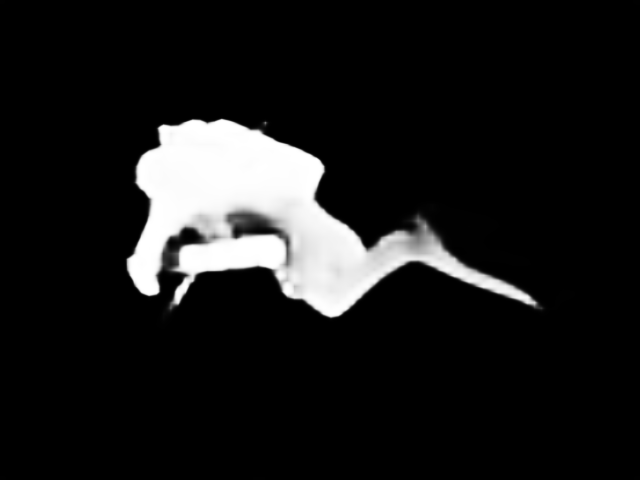} &
			\hspace{-0.46cm}
			\includegraphics[width = 0.082\linewidth]{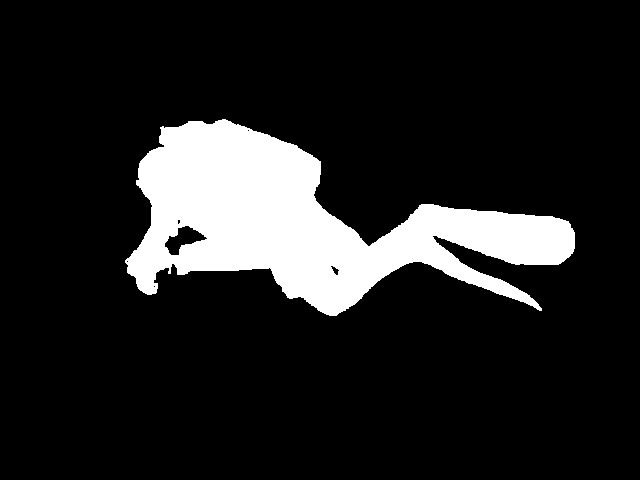} \\		
			
			\small Input
			& \hspace{-0.36cm} \small EUIVF
			& \hspace{-0.36cm} \small OCM
			& \hspace{-0.36cm} \small UDCP
			& \hspace{-0.36cm} \small TSA
			& \hspace{-0.36cm} \small UGAN
			& \hspace{-0.36cm} \small WaterNet
			& \hspace{-0.36cm} \small FGAN
			& \hspace{-0.36cm} \small UWCNN
			& \hspace{-0.36cm} \small Ours
			& \hspace{-0.36cm} \small GT
			
			\\
		\end{tabular}
	\end{center}
	\vspace{-4mm}
	\caption{\label{fig:saliency}Application examples of different methods on the salient object detection task in real-world USOD underwater dataset.}
	\vspace{-6mm}
\end{figure*}

%
To further verify that the improvement can be provided by our algorithm for high-level visual tasks, we apply an underwater salient object detection algorithm~\cite{f3net} to evaluate it on the benchmark USOD~\cite{islam_USOD} dataset.
Fig.~\ref{fig:saliency} shows that the saliency maps generated by our algorithm have a more integrated structure and precise boundary, even though in the dark underwater environment. By contrast, some state-of-the-art methods~\cite{Ancuti_EUIVF, Drews_UDCP, Fu_TSA} even fail to capture the rough outline of the objects.
We implement quantitative evaluation on the USOD dataset. As summarized in Table~\ref{tab:mae-usod}, the proposed algorithm performs favorably against the state-of-the-art methods in three common evaluation metrics (i.e., F-measure, S-measure, and MAE).
Application examples and quantitative results illustrate that the proposed underwater enhancement algorithm can make further effects on the implementation of relevant high-level vision tasks in the underwater environment.

\begin{table}[t]
	\centering
	\caption{Quantitative NIQE and UIQM values of different methods on real-world benchmark datasets.}\vspace{-2mm}
	\label{tab:niqe-uiqm}
	\renewcommand{\arraystretch}{1.2}
	\setlength{\tabcolsep}{1.82mm}{
		\begin{tabular}{c|l|l|l|l|l|l}
			\toprule
			\multirow{2}{*}{Methods} & \multicolumn{3}{c|}{NIQE $\downarrow$} & \multicolumn{3}{c}{UIQM $\uparrow$} \\ 
			\cline{2-7} 
			& UIEB  &  EUVP & REIU  & UIEB &   EUVP & REIU \\ 
			\hline
			\hline
			\multicolumn{1}{l|}{EUIVF~\cite{Ancuti_EUIVF}}    & $4.059$  &  $4.358$   &  $4.542$  &  $2.679$ & $2.763$  & $3.073$ \\ 
			\multicolumn{1}{l|}{OCM~\cite{Li_OCM}}      & $\textcolor{blue}{\textbf{3.877}}$  &  $4.628$   &  $4.538$  & $2.545$ & $2.776$  & $2.912$\\ 
			\multicolumn{1}{l|}{UDCP~\cite{Drews_UDCP}}     & $4.303$  &  $4.398$   &  $5.131$  & $1.772$ & $2.079$  & $2.099$\\
			\multicolumn{1}{l|}{TSA~\cite{Fu_TSA}}      & $4.165$  &  $5.623$   &  $4.842$  & $1.996$ & $2.869$  & $2.512$\\ 
			\multicolumn{1}{l|}{UGAN~\cite{Fabbri_UGAN}}    & $7.057$  &  $6.467$   &  $6.680$  & $2.528$ & $3.254$  & $3.043$\\ 
			\multicolumn{1}{l|}{WaterNet~\cite{Li_waternet}} & $4.484$  &  $4.375$   &  $4.544$  & $2.857$ & $3.065$  & $\textcolor{blue}{\textbf{3.150}}$\\ 
			\multicolumn{1}{l|}{AIO~\cite{Uplavikar_AIO}}      & $3.994$  &  $4.892$   &  $5.637$  & $3.078$ & $\textcolor{red}{\textbf{3.346}}$  & $3.137$\\  
			\multicolumn{1}{l|}{FGAN~\cite{Islam_FGAN}}    & $6.364$  &  $5.175$   &  $5.696$  & $2.512$ & $3.211$  & $2.982$\\ 
			\multicolumn{1}{l|}{UWCNN~\cite{Li_UWCNN}}    & $4.441$  &  $\textcolor{blue}{\textbf{4.251}}$   &  $\textcolor{blue}{\textbf{4.411}}$  & $\textcolor{blue}{\textbf{3.078}}$ & $2.231$  & $2.781$\\ 
			\multicolumn{1}{l|}{Ucolor~\cite{Li_Ucolor}}   & $3.772$  &  $--$   &  $--$  & $2.871$ & $--$  & $--$\\ 
			\multicolumn{1}{l|}{Ours}     & $\textcolor{red}{\textbf{3.451}}$  &  $\textcolor{red}{\textbf{4.165}}$   &  $\textcolor{red}{\textbf{3.694}}$  & $\textcolor{red}{\textbf{3.763}}$ & $\textcolor{blue}{\textbf{3.297}}$  & $\textcolor{red}{\textbf{3.966}}$\\ 
			\bottomrule
	\end{tabular}}
\vspace{-1mm}
\end{table}

\begin{table*}[t]
	\footnotesize
	\caption{Quantitative comparisons on salient object detection task with state-of-the-art underwater enhancement methods.}
	\label{tab:mae-usod}
	\vspace{-3mm}
	\centering
	\setlength{\tabcolsep}{1.50mm}{
		\renewcommand{\arraystretch}{1.2}
		\begin{tabular}{l|ccccccccc|c}
			\toprule
			Metric & EUIVF~\cite{Ancuti_EUIVF} & OCM~\cite{Li_OCM} & UDCP~\cite{Drews_UDCP} & TSA~\cite{Fu_TSA} & UGAN~\cite{Fabbri_UGAN} & WaterNet~\cite{Li_waternet} & AIO~\cite{Uplavikar_AIO} & FGAN~\cite{Islam_FGAN} & UWCNN~\cite{Li_UWCNN} &  Ours\\
			\hline
			F-measure $\uparrow$ & $0.850$ & $0.840$ & $0.835$ & $0.715$ & $0.836$ &  $0.852$ & $0.813$ & $\textcolor{blue}{\textbf{0.851}}$ & $0.834$ &  $\textcolor{red}{\textbf{0.854}}$\\
			S-measure $\uparrow$ & $\textcolor{blue}{\textbf{0.833}}$ & $0.831$ &  $0.811$ & $0.705$ & $0.822$ & $\textcolor{blue}{\textbf{0.833}}$ & $0.797$ & $0.830$ & $0.808$ &  $\textcolor{red}{\textbf{0.837}}$\\
			MAE $\downarrow$ & $\textcolor{blue}{\textbf{0.081}}$ & $\textcolor{blue}{\textbf{0.081}}$ & $0.088$ & $0.124$ & $0.084$ & $\textcolor{red}{\textbf{0.080}}$ & $0.096$ & $0.082$ & $0.092$ &  $\textcolor{red}{\textbf{0.080}}$\\
			\bottomrule
	\end{tabular}}
	\vspace{-2mm}
\end{table*}

\begin{table}[t]
	\vspace{-2mm}
	\footnotesize
	\caption{Ablation analysis on the UIEB dataset.}
	\label{tab:ablation}
	\vspace{-2mm}
	\centering
	\renewcommand{\arraystretch}{1.0}
	\setlength{\tabcolsep}{1.72mm}{
		\begin{tabular}{l|c|ccc|c|cc}
			\toprule
			\multirow{2}{*}{\textbf{Method}} &\multirow{2}{*}{\textbf{BaseNet}} &\multicolumn{3}{c|}{\textbf{SFANet}} & \multirow{2}{*}{\textbf{FDNet}}& \multicolumn{2}{c}{\textbf{UIEB}}\\
			&   &$F$ & $\mathcal{B}_{st}$ & $\mathcal{B}_{te}$ && PSNR & SSIM\\
			\specialrule{0em}{1pt}{1pt}
			\hline
			\specialrule{0em}{1pt}{1pt}
			${M}_{0}$ & \checkmark &  &  &    && $20.244$ & $0.8772$\\
			${M}_{1}$ & \checkmark & \checkmark &  &   && $20.943$ & $0.8932$\\
			${M}_{2}$ & \checkmark &  & \checkmark &   &\checkmark& $20.879$ & $0.8821$\\
			${M}_{3}$ & \checkmark &  &  & \checkmark  &\checkmark& $22.202$ & $0.8893$\\
			${M}_{4}$ & \checkmark   &  & \checkmark & \checkmark & & $21.846$ & $0.8918$\\
			Ours      & \checkmark  &  & \checkmark & \checkmark  & \checkmark& $\textbf{22.448}$ & $\textbf{0.9019}$\\
			\bottomrule
	\end{tabular}}
	\vspace{-3.2mm}
\end{table}

\subsection{Ablation Studies}
\vspace{-1mm}
In Table~\ref{tab:ablation}, ${M}_{0}, {M}_{1}, \ldots, {M}_{3}$ refer to algorithms implemented for ablation analysis.
${M}_{0}$ is the UNet-like base network called BaseNet for underwater image enhancement. 
${M}_{1}$ refers to that the multi-scale features $F$ from the semantic-aware Feature Aggregation network (SFANet) are directly embedded into the decoder of the BaseNet. 
$M_{2}$ is a variant of our algorithm with only structure branch $\mathcal{B}_{st}$ in the SFANet, and $M_{3}$ is another variant with only texture branch $\mathcal{B}_{te}$.
$M_{4}$ denotes that the final aggregated features of the SFANet are directly embedded into the decoder of the base network without the feature dominative network (FDNet).

\begin{figure}[t]
	\begin{center}
		\begin{tabular}{ccc}
			\includegraphics[width = 0.251\linewidth]{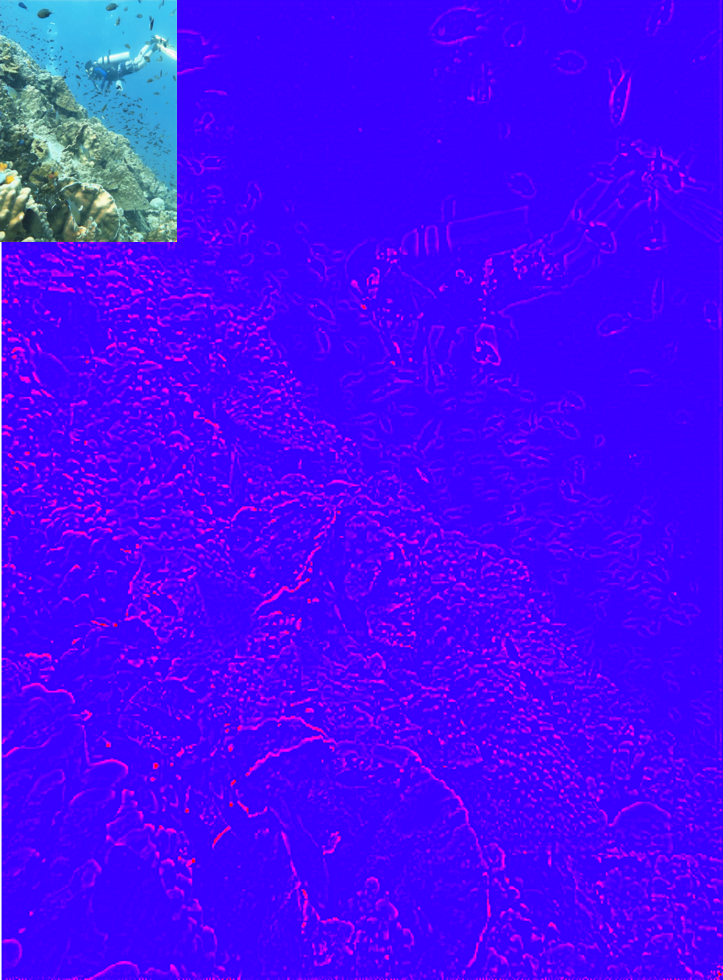}   & \hspace{-0.46cm}
			\includegraphics[width = 0.25\linewidth]{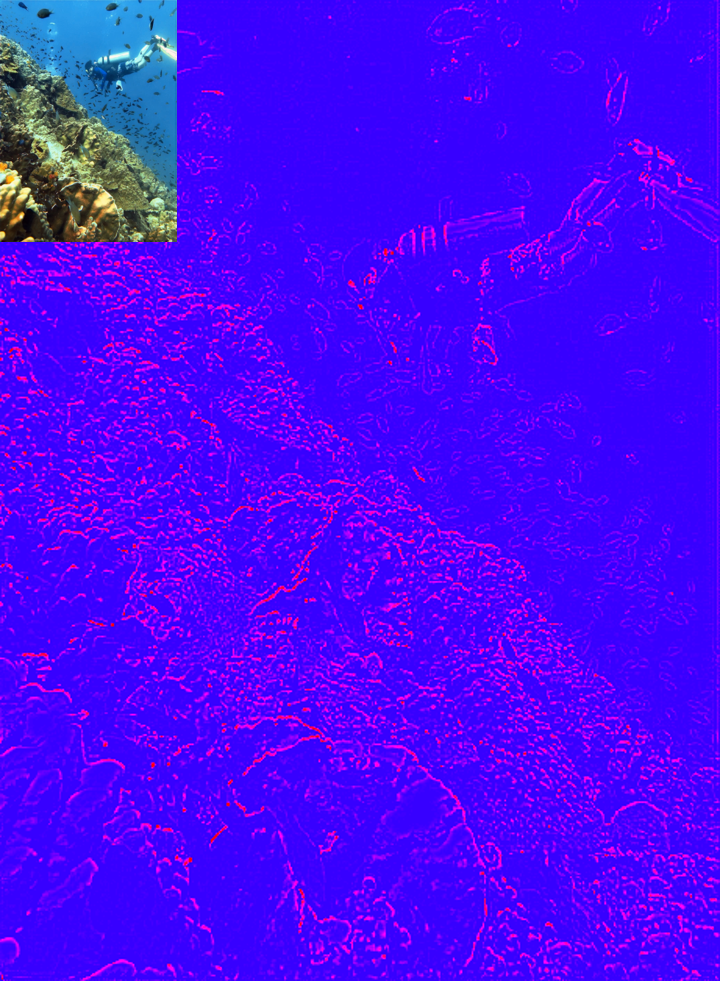} & \hspace{-0.56cm}
			\includegraphics[width = 0.42\linewidth]{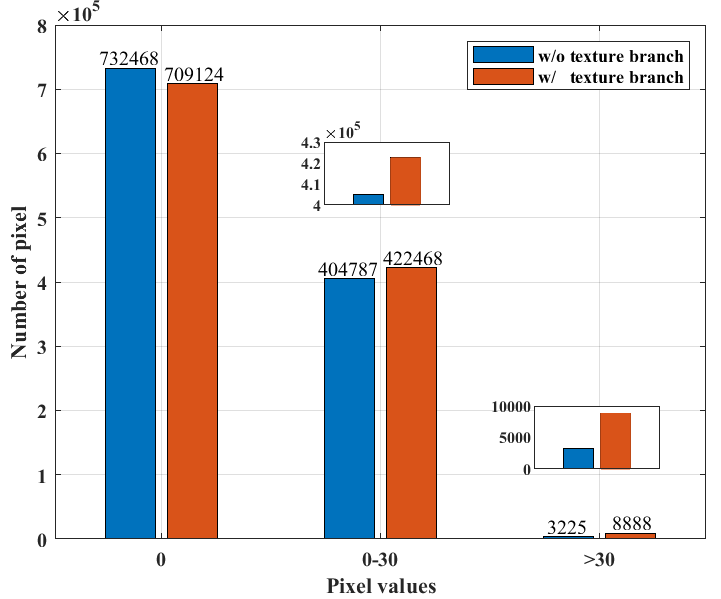} \\					
			
			(a) 
			& \hspace{-0.36cm} (b) 
			& \hspace{-0.36cm} (c) \\
		\end{tabular}
	\end{center}
	\vspace{-4mm}
	\caption{\label{fig:tex_map}Ablation analysis of the texture branch. (a) Texture layer of the enhanced image of the algorithm without $\mathcal{B}_{te}$. (b) Texture layer of the enhanced image of the algorithm with $\mathcal{B}_{te}$. (c) Histogram comparison of (a) and (b).}
	\vspace{-6mm}
\end{figure}

\begin{figure}[t]
	\vspace{-2mm}
	\begin{center}
		\begin{tabular}{cc}
			\includegraphics[width = 0.47\linewidth]{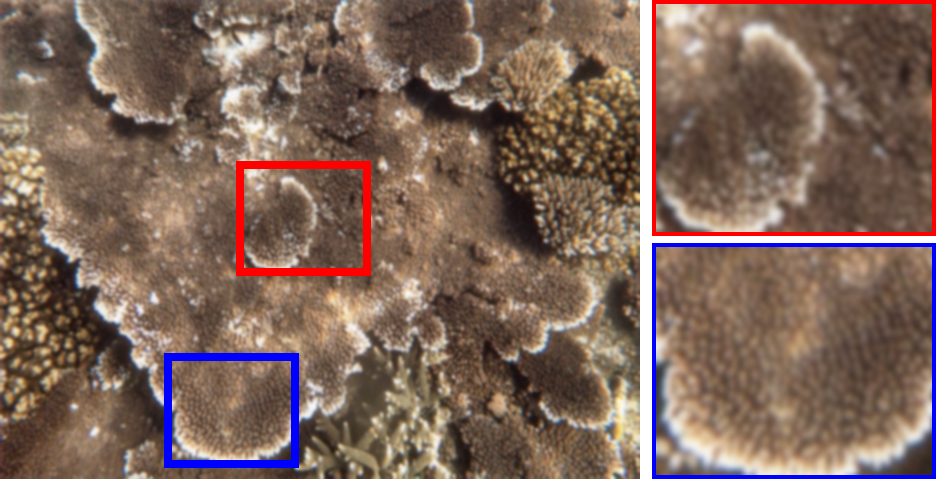}   & \hspace{-0.46cm}
			\includegraphics[width = 0.47\linewidth]{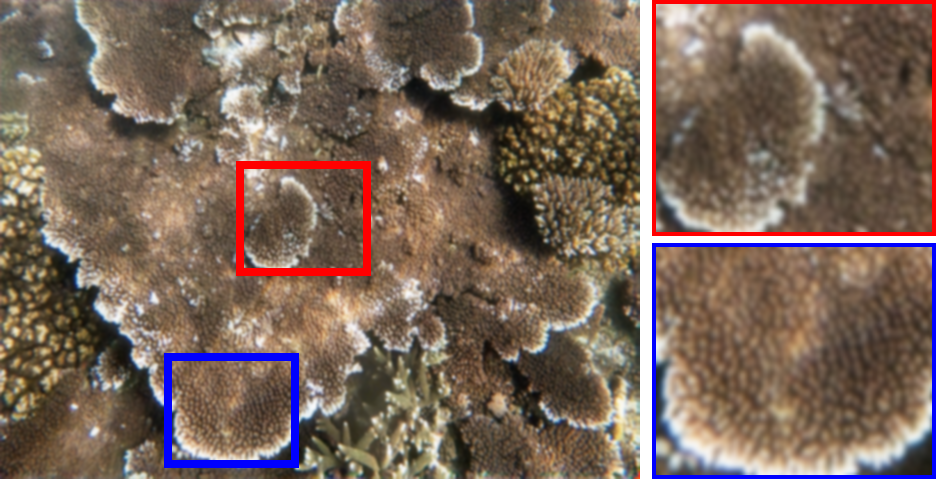} \\					
			(a) w/o structure branch 
			& \hspace{-0.36cm} (b) w/ structure branch \\
		\end{tabular}
	\end{center}
	\vspace{-4mm}
	\caption{\label{fig:st_map}Ablation analysis of the structure branch. (a) and (b) are structure layers through gaussian filtering corresponding to the enhanced results without and with structure branch, respectively.}
	\vspace{-2mm}
\end{figure}

\textbf{Effectiveness of Texture and Structure Branches.}
%
We can observe from Table~\ref{tab:ablation} that: 
i)
On the whole, compared with ${M}_{0}$, ${M}_{1}$ only obtains a gain of $0.70$dB, while our algorithm outperforms ${M}_{1}$ by a large margin \textbf{(1.51dB)}. It can be explained that texture and structure branches are indeed conducive to maximizing feature utilization for underwater image enhancement.
%
ii) For texture branch $\mathcal{B}_{te}$, the performance of the algorithm with it is improved by $1.96$dB via comparing $M_{3}$ with ${M}_{0}$. 
To more explicitly demonstrate the effect of texture branch, a Gaussian filter is applied to the enhanced images to obtain texture layers for analysis. 
As shown in Fig.~\ref{fig:tex_map}(a) and (b), the enhanced image of the algorithm with $\mathcal{B}_{te}$ has more textures. Besides, the statistical results in Fig.~\ref{fig:tex_map}(c) show that the texture branch contributes to the restoration of image textures. 
iii) For structure branch $\mathcal{B}_{st}$, the algorithm with it gains $0.64$dB improvement compared with ${M}_{0}$. The corresponding qualitative results in Fig.~\ref{fig:st_map} show the content of the structure layer enhanced by the algorithm without $\mathcal{B}_{st}$ is seriously blurred.
Therefore, we argue that texture and structure branches can model the correlation between features, and both of them are significant to generate near-lossless underwater image content.

\textbf{Effectiveness of Feature Dominative Network.} 
To illustrate the effectiveness of the feature dominative network (FDNet), we also train the model without the feature dominating process. As shown in Table~\ref{tab:ablation}, using the FDNet can obtain a gain of $0.58$dB, which indicates that the FDNet can reasonably dominate aggregated features from the SFANet to make them more applicable to underwater image enhancement.

\textbf{Effectiveness of Multi-path CFRM.}
To show the effectiveness of the multi-path Contextual Feature Refinement Module (CFRM), we remove it as our comparison model.
As shown in Table~\ref{tab:ab-SFTM}, an improvement of $0.879$dB is obtained using CFRM, suggesting that the feature refinement module can highlight more informative features by modeling the correlation between features.

\begin{table}[t]
	\footnotesize
	\caption{Investigation of the multi-path CFRM on the UIEB dataset.}
	\label{tab:ab-SFTM}
	\vspace{-2mm}
	\centering
	\setlength{\tabcolsep}{4.0mm}{
		\begin{tabular}{lcc}
			\toprule
			& \textbf{w/o CFRM} &\textbf{ w/ CFRM}\\
			\hline
			\specialrule{0em}{1pt}{1pt}
			PSNR & $21.569$ & $22.448 _{\color{red}{\uparrow{0.879}}}$\\
			SSIM & $0.8902$ & $0.9019 _{\color{red}{\uparrow{0.012}}}$\\
			\bottomrule
	\end{tabular}}
	\vspace{-6mm}
\end{table}

\subsection{Additional Analysis}
\textbf{Encoder embedding \textit{vs.} Decoder embedding.}
We conduct feature embedding investigations on the encoder and decoder of the UNet-like base network, respectively.
As shown in Table~\ref{tab:posi-embed}, we note that the performance of the proposed algorithm is improved by a large margin by embedding the aggregated features from the SFANet into the decoder rather than the encoder, on the UIEB and EUVP datasets.
It indicates that embedding features from the semantic-aware model into the decoder of the base network for feature reconstruction can maximize feature utilization and achieve better enhancement performance.

\begin{table}[h]
	\vspace{-2mm}
	\footnotesize
	\caption{Analysis of the feature embedding location.}
	\label{tab:posi-embed}
	\vspace{-3mm}
	\centering
	\setlength{\tabcolsep}{1.5mm}{
		\begin{tabular}{lccccc}
			\toprule
			\multirow{2}{*}{\textbf{Dataset}} &\multicolumn{2}{c}{\textbf{Encoder}} &&\multicolumn{2}{c}{\textbf{Decoder}}\\
			\specialrule{0em}{1pt}{1pt}
			& PNSR & SSIM && PNSR & SSIM\\
			\hline
			\specialrule{0em}{1pt}{1pt}
			UIEB~\cite{Li_waternet} & $21.703$  & $0.8836$ && $22.448 _{\color{red}{\uparrow{0.745}}}$ & $0.9019$ \\
			EUVP~\cite{Islam_FGAN} & $22.858$  & $0.9007$ && $23.079 _{\color{red}{\uparrow{0.221}}}$ & $0.9033$\\
			\bottomrule
	\end{tabular}}
	\vspace{-4mm}
\end{table}

\section{CONCLUSIONS}
In this paper, we proposed an efficient underwater image enhancement algorithm based on semantic-aware texture and structure feature collaboration. 
%
%
The main novel points of the proposed algorithm lie in that the hierarchical features of the high-level semantic-aware model were exploited as an auxiliary and were characterized via structure and texture branches. A multi-path contextual feature refinement module was deployed on both branches to model the correlation between features resulting in near-lossless image content.
%
%
%
We presented a feature dominative network to perform channel-wise feature modulation to adapt to the different feature patterns of the enhancement network.
Experiments implemented on four widely-used underwater benchmarks demonstrated the superiority of our algorithm and its semantic-aware ability for high-level vision tasks.
In future work, we plan to explore the potential of the proposed algorithm in the field of underwater machine vision, such as underwater robotics and underwater object detection.
\addtolength{\textheight}{-3.0cm}   







\newpage
\Large
\bibliographystyle{IEEEtranS}
\bibliography{IEEEexample}

\end{document}